%% file: main.tex
\documentclass{article}

\usepackage{microtype}
\usepackage{graphicx}
\usepackage{subcaption}
\usepackage{booktabs}
\usepackage{hyperref}
\usepackage{dblfloatfix}
\usepackage{float}

\usepackage[accepted]{icml2026}

\makeatletter
\renewcommand{\ICML@appearing}{\textit{Spotlight paper at the ICML 2026 Workshop
on Mechanistic Interpretability}. Copyright 2026 by the author(s).}
\makeatother

\usepackage{amsmath}
\usepackage{amssymb}
\usepackage{mathtools}
\usepackage{amsthm}
\usepackage{enumitem}
\usepackage[capitalize,noabbrev]{cleveref}
\usepackage{xcolor}
\usepackage{alphalph}

\makeatletter
\input{custom_commands}
\makeatother
\draftcommentsfalse

\newif\ifdiagprovenance
\diagprovenancetrue
\newcommand{\DIAG}[2]{%
	\ifdiagprovenance\,{\scriptsize\textcolor{gray!70}{[\repofile[#1]{#2}]}}\fi%
}

\setlist[itemize]{leftmargin=*,topsep=2pt,itemsep=1pt,parsep=0pt}

\icmltitlerunning{Size Doesn't Matter: Cosine-Scored Sparse Autoencoders}

\newif\ifshownoaux
\shownoauxtrue

\newif\ifshoworphan
\showorphanfalse

\begin{document}

\twocolumn[
\icmltitle{Size Doesn't Matter: Cosine-Scored Sparse Autoencoders}

\begin{icmlauthorlist}
	\icmlauthor{Silen Naihin}{exp}
	\icmlauthor{Lev Stambler}{tear}
\end{icmlauthorlist}
\icmlaffiliation{exp}{Experiential Labs}
\icmlaffiliation{tear}{Tear Labs}
\icmlcorrespondingauthor{Silen Naihin}{silen@experientiallabs.ai}

\icmlkeywords{Sparse Autoencoders, Mechanistic Interpretability, Dictionary Learning}

\vskip 0.3in
]

\printAffiliationsAndNotice{}

\begin{abstract}
Sparse autoencoders (SAEs) detect features via inner product, so a feature's activation scales with both its directional alignment and the input's norm.
Features that fire on token norm therefore claim dictionary slots regardless of content alignment. This matters because sublayer normalization has already discarded the magnitude the score measures, so the encoder detects a quantity the model does not read.
We replace the score with a learned blend of cosine similarity and input magnitude, letting the optimizer choose how much norm to use; a per-feature extension lets each feature decide independently. In both regimes, training is free to recover inner product but never does, with no feature ever choosing more than half-magnitude dependence.
At matched reconstruction, the cosine encoder learns features that align with human-recognizable concepts far more often than standard, filling dictionary slots that inner product wastes on norm detectors. Loss reweighting that equalizes gradients barely closes the gap, confirming forward-pass score geometry as the lever.
The advantage is not universal across tasks or depths, but we believe cosine scoring should be the default for dictionary learning on normalized representations.
\end{abstract}

\input{sections/body/01-introduction}
\input{sections/body/02-background}
\input{sections/body/03-method}
\input{sections/body/04-experiments}
\input{sections/body/05-limitations}
\input{sections/body/06-discussion}

\section*{Impact Statement}
This paper advances mechanistic interpretability and dictionary learning. We do not identify additional societal consequences beyond those generally associated with interpretability research.

\bibliography{refs}
\bibliographystyle{icml2026}

\newpage
\onecolumn
\appendix
\renewcommand{\thesection}{\AlphAlph{\value{section}}}
\input{sections/03-related-work}
\input{sections/04-cosine-sae-architecture}
\input{sections/05-diagnostic-evidence}
\input{sections/06-production-results}
\input{sections/06b-sample-efficiency}
\input{sections/06c-interpretability}
\input{sections/07-gradient-equalization}
\input{sections/08-limitations}

\input{sections/08b-recommendations}
\input{sections/09-discussion}
\input{sections/appendix}

\end{document}

%% file: custom_commands.tex
\newif\ifdraftcomments
\draftcommentstrue   

\definecolor{levcolor}{HTML}{1F77B4}    
\definecolor{silencolor}{HTML}{2CA02C}  
\definecolor{codexcolor}{HTML}{D62728}  
\definecolor{claudecolor}{HTML}{9467BD} 

\newcommand{\@feedbackcomment}[3]{%
  \ifdraftcomments
    \begingroup
      \small\color{#1}%
      \textbf{[#2:}\,{\sffamily #3}\textbf{]}%
    \endgroup
  \fi%
}

\newcommand{\lev}[1]{\@feedbackcomment{levcolor}{lev}{#1}}
\newcommand{\silen}[1]{\@feedbackcomment{silencolor}{silen}{#1}}
\newcommand{\codex}[1]{\@feedbackcomment{codexcolor}{codex}{#1}}
\newcommand{\claude}[1]{\@feedbackcomment{claudecolor}{claude}{#1}}
\newcommand{\levForAI}[1]{\@feedbackcomment{levcolor}{lev says for AI to do this:}{#1}}
\newcommand{\QlevForAI}[1]{\@feedbackcomment{levcolor}{Question lev has for AI:}{#1}}

\newcommand{\repourl}{https://github.com/SilenNaihin/cosine-scored-saes}
\newcommand{\repofileurl}{\repourl/blob/main}
\newcommand{\repofile}[2][]{%
  \def\repofile@label{#1}%
  \ifx\repofile@label\@empty
    \href{\repofileurl/#2}{\texttt{#2}}%
  \else
    \href{\repofileurl/#2}{#1}%
  \fi
}

%% file: sections/body/01-introduction.tex
\section{Introduction}
\label{sec:intro-short}
\label{sec:intro}

Sparse autoencoders are a standard dictionary-learning tool for mechanistic interpretability \cite{bricken2023monosemanticity,cunningham2023sparse,gao2024scaling,bussmann2024batchtopk,karvonen2025saebench}.
A trained SAE is read as a feature dictionary: if feature $i$ fires on activation $x$, then $x$ is taken to contain the concept represented by the corresponding decoder direction.
The interpretation depends on how the encoder detects those directions. The standard rule is an inner product $\langle w_i, x\rangle$, so a feature fires in proportion to both its directional alignment with $x$ and the magnitude $\|x\|$.

\begin{figure*}[!t]
  \centering
  \includegraphics[width=\textwidth]{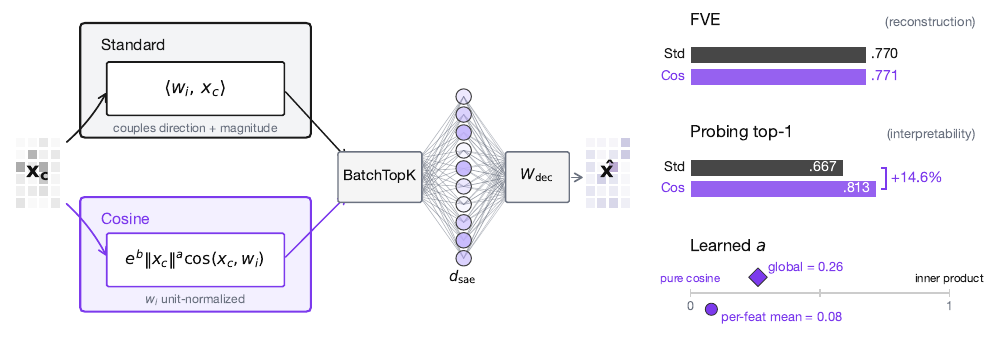}
  \caption{\textbf{The cosine encoder: architecture and headline results.}
    \textbf{Left:} Standard SAE encoder computes $\langle w_i, x_c\rangle$, coupling alignment with norm. Cosine encoder unit-normalizes $w_i$ and replaces the score with $e^b\|x_c\|^a\cos(x_c, w_i) + b_{\mathrm{enc},i}$; $a$ interpolates between pure cosine ($a{=}0$) and inner product ($a{=}1$).
    \textbf{Right:} Matched reconstruction (FVE $\approx 0.77$) with $+14.6$\% sparse-probing top-1 (mean over three SAE-training seeds). Training drives $a \approx 0.26$, far from the inner-product limit.
    Qwen3-8B L18, 500M tokens, $d_{\mathrm{sae}}{=}65{,}536$.}
  \label{fig:architecture}
\end{figure*}

\begin{figure*}[t]
  \centering
  \includegraphics[width=\textwidth]{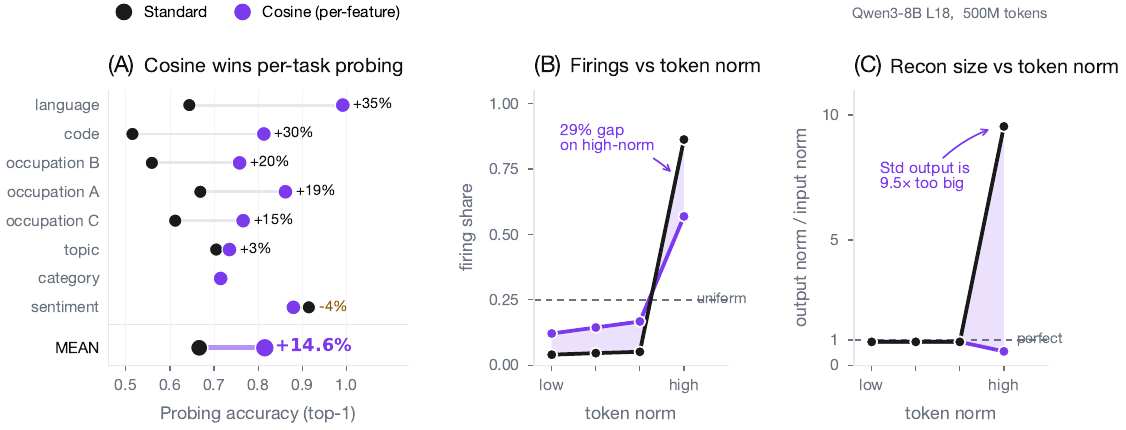}
  \caption{\textbf{Cosine-scored SAEs win on probing because standard features fire on token norm.}
    \textbf{(A)} Result: sparse-probing top-1 across eight tasks (Qwen3-8B L18, 500M tokens, $d_{\mathrm{sae}}{=}65{,}536$, matched FVE $\approx 0.77$). Per-feature cosine wins on 7/8 tasks; sentiment is the only exception.
    \textbf{(B)} Cause: standard's \emph{unmatched} features (those with no nearest-neighbor counterpart in cosine's dictionary) fire $22\times$ more on the highest-norm token quartile than on the lowest, versus $4.7\times$ for cosine; they encode magnitude, not content.
    \textbf{(C)} Confirmation: on the same high-norm tokens, the standard SAE reconstructs at $9.5\times$ the input norm, while cosine stays close to the input scale ($0.55\times$). Removing the $\|x\|$ factor from the score recovers content-encoding features.
    Provenance: Appendix~\ref{app:saebench}.}
  \label{fig:hero}
\end{figure*}

\begin{figure*}[t]
  \centering
  \includegraphics[width=\textwidth]{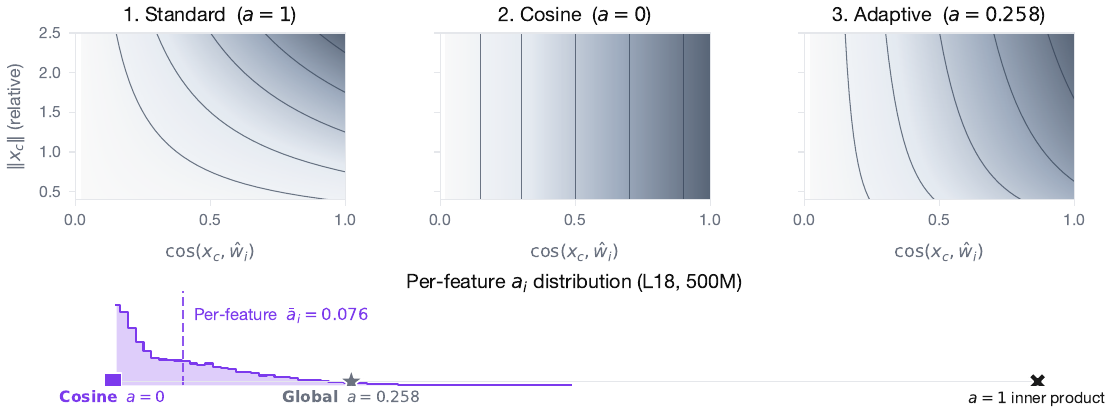}
  \caption{\textbf{Score-surface geometry.} Each panel plots the encoder pre-activation $\|x_c\|^a\cos(x_c, w_i)$ over alignment (x-axis) and input norm (y-axis). Black curves join equally-scored pairs. \emph{Left:} $a{=}1$ (inner product); hyperbolic curves, high-norm tokens outscore better-aligned low-norm ones. \emph{Center:} $a{=}0$ (cosine); vertical curves, norm ignored. \emph{Right:} global learned $a{\approx}0.26$; mild tilt, close to cosine. Bottom: per-feature $a_i$ distribution at the headline setting.}
  \label{fig:arch-overview}
\end{figure*}

This is the wrong scoring geometry for transformers with pre-sublayer normalization.
The inner-product score mixes content alignment with token norm, but normalization strips that norm before the model reads the activation; the encoder detects a quantity the downstream computation ignores.
This mismatch compounds under TopK selection \cite{bussmann2024batchtopk}: features compete for a limited budget by score, so norm-inflated features crowd out better-aligned ones regardless of content. BatchTopK's batch-wide budget sharpens this, with high-norm tokens claiming slots across the batch, but the crowding is a property of the score, not the selector, and persists when each token is scored independently.
Over training, this selection pressure starves content-encoding features: they rarely win TopK slots, receive no learning signal, and never develop.
The result is a dictionary dominated by features that fire on magnitude rather than meaning.
This is not a minor calibration issue. If dictionary elements fire on norm rather than content, they cannot reliably be used for sparse probing, feature-based steering, or circuit analysis; the practitioner interprets them as model concepts, but the features encode a quantity the model discards.
A score aligned with what the model actually reads should depend on direction, not magnitude.

We replace the inner-product score with a cosine score scaled by a learned exponent $a$ on the input norm:
\[
  s_i(x) \;=\; \|x\|^a \cdot \cos(x, w_i) \;+\; b_{\mathrm{enc},i},
\]
which interpolates between cosine ($a=0$) and inner product ($a=1$); we use unit-normalized encoder rows so $a{=}0$ recovers cosine cleanly (\S\ref{sec:method}).\footnote{Up to a learned global temperature; see \S\ref{sec:method-global}.}
We study two parameter regimes: a single global $a$ shared across features, and a per-feature extension $a_i = a_{\mathrm{base}} + \delta_i$ that lets each feature learn its own norm dependence.
The key property is that the optimizer is free to recover inner product ($a{=}1$) if magnitude is genuinely useful, but is not forced to use it by default. This lets training discover the right geometry rather than assuming it.
In both regimes, training consistently drives $a$ toward zero; no feature ever approaches the inner-product regime, confirming that direction, not magnitude, is the useful signal.

Through experiments on Qwen3-8B and Gemma-2-2B, we demonstrate that cosine-scored SAEs match standard BatchTopK on reconstruction fidelity, model-behavior preservation, and per-feature interpretability, while producing dictionaries that align far more often with human-recognizable concepts. A matched-feature decomposition reveals that the majority of the probing gap comes not from improving shared features, but from features the standard encoder never learns at all; its dictionary slots are instead occupied by norm detectors. We isolate the cause to forward-pass TopK selection under norm inflation: loss reweighting that equalizes gradients across norm levels barely closes the gap, confirming that the score geometry itself, not the training signal, is the lever. The advantage replicates across layers and models, though it is not universal; deep LayerNorm layers and sentiment are exceptions where magnitude carries task-relevant signal.

\textbf{Contributions.}
\begin{itemize}[leftmargin=1.2em,itemsep=2pt,topsep=2pt]
  \item \textbf{Architecture.} We introduce cosine-scored sparse autoencoders, replacing the inner-product encoder score with a learned-exponent cosine score that interpolates between pure cosine ($a{=}0$) and inner product ($a{=}1$). We study global and per-feature parameterizations and identify the design choices that make each stable (\S\ref{sec:method}).

  \item \textbf{Result.} At matched reconstruction, KL, and per-feature interpretability, the cosine encoder lifts sparse-probing top-1 by $+14.6$\% on Qwen3-8B. A matched-feature decomposition shows that the majority of the gap comes from features the standard encoder never learns at all, rather than from improving shared features. The result replicates across layers and on Gemma-2-2B (\S\ref{sec:exp-headline}, \S\ref{sec:exp-mechanism}).

  \item \textbf{Mechanism.} The standard encoder's unmatched features fire overwhelmingly on high-norm tokens; they encode magnitude, not content. Forward-pass TopK selection under norm inflation is the primary cause; loss reweighting that equalizes gradients across norm levels barely closes the gap, confirming that the score geometry itself is the lever. The optimizer independently confirms this: training drives $a$ far from the inner-product regime, with no feature ever choosing more than half-magnitude dependence (\S\ref{sec:exp-mechanism}, \S\ref{sec:exp-grad}).
\end{itemize}

\noindent For practitioners, this is a drop-in encoder change that produces more interpretable dictionaries at no reconstruction cost. More broadly, it suggests that scoring geometry is an underexplored design axis for dictionary learning: if inner product is the wrong inductive bias for normalized representations, standard SAE dictionaries trained on modern transformers may systematically undercount content features. We believe cosine scoring should be the default for dictionary learning on normalized representations, and that the failure mode we identify likely affects any inner-product SAE trained on a post-RMSNorm site.

%% file: sections/body/02-background.tex
\section{Background}
\label{sec:background}
\label{sec:related}

A sparse autoencoder (SAE) takes an activation $x \in \mathbb{R}^{d}$, projects it through a single-layer encoder--decoder pair to a sparse code $z \in \mathbb{R}^{d_{\mathrm{sae}}}$, and reads the decoder rows $\{W_{\mathrm{dec},i}\}$ as a feature dictionary:
\[
    z      = \sigma\!\left((x - b_{\mathrm{dec}}) W_{\mathrm{enc}}^{\!\top} + b_{\mathrm{enc}}\right) \;
    \hat x = z W_{\mathrm{dec}} + b_{\mathrm{dec}}.
\]
with reconstruction loss: 
\(
    \mathcal{L} = \|x - \hat x\|^2.
\)
Different SAE variants make $\sigma$ sparse in different ways, including TopK \cite{gao2024scaling}, BatchTopK \cite{bussmann2024batchtopk}, JumpReLU \cite{rajamanoharan2024jumprelu}, gated activations \cite{rajamanoharan2024gated}, and AbsTopK \cite{zhu2025abstopk}.
SAEs tend to assume the linear representation hypothesis: concepts correspond to linear directions in activation space \cite{park2024linear,park2025geometry,elhage2022superposition}.
Recent variants modify dictionary geometry \cite{korznikov2025ortsae,bussmann2025matryoshka} or loss shape \cite{nasiri2026monoloss} but leave the encoder score unchanged. A feature's pre-activation is still an inner product $\langle w_i, x\rangle = \|w_i\|\,\|x\|\,\cos(w_i, x)$, so larger-norm tokens raise the score even when directional alignment is unchanged.
Following standard practice \cite{bricken2023monosemanticity,gao2024scaling,karvonen2025saebench}, decoder rows are constrained to unit norm during training: re-normalized after each optimizer step, with the component of the decoder gradient parallel to each row projected away to account for the Adam-normalization interaction \cite{gao2024scaling}. The dictionary therefore represents directions, and any per-feature scale is absorbed into the encoder. The encoder reads the input centered on the decoder bias ($x \mapsto x - b_{\mathrm{dec}}$, as in the equation above), tying its zero point to the reconstruction origin \cite{bricken2023monosemanticity}.

We hook the SAE on the residual stream: the activation $x$ that flows untouched through every transformer block, with each sublayer adding to it.
Most modern transformers place an RMSNorm \cite{zhang2019rmsnorm} on the path \emph{into} every sublayer:
\[
  \mathrm{RMSNorm}(x) \;=\; \sqrt{d}\,\frac{x}{\|x\|} \odot g, \qquad g \in \mathbb{R}^{d}.
\]
Each sublayer therefore reads $x/\|x\|$ up to the per-coordinate gain $g$, not $x$ itself.
The SAE, hooked one step earlier on the residual stream, still sees $x$ in full.
Two activations with the same direction and different norms look nearly identical to the model but very different to an inner-product encoder, which scores them in the ratio of their norms.
Residual-stream norms are heavy-tailed in practice, driven by rogue dimensions \cite{timkey2021rogue} and outlier features \cite{dettmers2022llmint8}, so this bias is not hypothetical.

Directly swapping activation directions between random token pairs (no SAE involvement) produces $219\times$ more downstream KL-divergence than swapping their norms at the same layer (Appendix~\ref{app:direction-patching}). The model's computation is direction-dominated; the SAE's encoder should be too.

We use BatchTopK throughout \cite{bussmann2024batchtopk}, the SAEBench default \cite{karvonen2025saebench}. BatchTopK imposes a single batch-wide activation budget rather than forcing exactly $k$ features per token, letting the model allocate more features to complex tokens and fewer to simple ones. Given pre-activations $u = \sigma(x W_{\mathrm{enc}}^{\!\top} + b_{\mathrm{enc}}) \in \mathbb{R}^{N \times d_{\mathrm{sae}}}$ over a batch of $N$ tokens, $z = \operatorname{BatchTopK}(u)$ keeps only the $kN$ largest entries and zeros the rest. This flexibility also means high-norm tokens can claim disproportionately many slots: their inflated pre-activations dominate the batch-wide ranking regardless of directional alignment. Batch-wide competition is one route by which magnitude corrupts selection, but it is not the only one; the inner-product score conflates direction with magnitude through the input norm and the encoder-row norm alike, so the pathology survives even when each token is scored independently. Our cosine score modifies only the encoder pre-activation, not the sparsity mechanism, so it composes with any selector, and we confirm the advantage holds under per-token TopK and AbsTopK as well as BatchTopK (our default; Appendix~\ref{app:selectors}). Penalty-trained selectors (JumpReLU, gated) are left to future work.

More broadly, modern architectures are adding normalization (QK-norm \cite{henry2020qknorm,dehghani2023scaling22b}, nGPT \cite{loshchilov2024ngpt}), strengthening the case for direction-aware SAE encoders.

\textbf{Evaluation metrics.}
\label{sec:metrics}
We follow SAEBench \cite{karvonen2025saebench} throughout.
Reconstruction is reported as fraction of variance explained, $\mathrm{FVE} = 1 - \mathbb{E}\|x - \hat x\|^2 / \mathrm{Var}(x)$, computed per token and aggregated.
Sparse probing fits a linear probe restricted to the $k$ SAE features that best predict a labeled concept and reports the resulting accuracy (top-$k$) over eight classification datasets; top-1 isolates the single best feature and is the most direct test of whether one dictionary direction matches a human-recognizable concept.
Auto-interp asks an LLM to describe a feature from its top-activating contexts and a second LLM to score whether the description predicts the firing pattern; we report the fraction of features judged interpretable (Appendix~\ref{sec:interpretability}).

%% file: sections/body/03-method.tex
\section{The Cosine Encoder}
\label{sec:method}
\label{sec:arch-short}

The fix is straightforward in principle: remove $\|x\|$ from the score. The design question is how much norm information to retain, since the decoder must still reconstruct $x$ at full scale. We define a one-parameter family that lets the optimizer answer this question.

\textbf{Notation.} Let $x_c = x - b_{\mathrm{dec}}$ be the input centered on the decoder bias. We keep encoder rows $w_i$ unit-normalized, recomputing $w_i \leftarrow w_i/\|w_i\|$ on every forward pass with gradients flowing through the normalization.\footnote{So cosine is just an inner product on the sphere: with $\|w_i\|=1$, $\cos(x_c, w_i) = \langle x_c/\|x_c\|, w_i \rangle$, and at $a{=}1$ the score below reduces to $\langle x_c, w_i\rangle$ exactly.} Decoder rows are also unit-normalized, per the held-fixed community recipe. Encoder unit-normalization is what makes the score true cosine. Dropping it (keeping input-side normalization but using raw $w_i$) yields $\|w_i\|\cdot\cos(x_c, w_i)$, a half-cosine that reintroduces $\|w_i\|$ as a magnitude factor. $\|w_i\|$ then drifts unconstrained (no loss term penalizes it) and the dictionary collapses to $93$\% dead features at L27/50M.\footnote{The full design-space matrix (these three settings plus per-axis ablations on encoder/decoder normalization and post-decode norm restoration) is in Appendix~\ref{sec:ablation-matrix}.}

\textbf{Score, pipeline, and initialization.} We define the pre-activation as
\[
  s_i(x) \;=\; \underbrace{\exp\!\big(a \log\|x_c\| + b\big)}_{e^b\,\|x_c\|^a} \cdot \cos(x_c, w_i) \;+\; b_{\mathrm{enc}, i},
\]
The log-exp parameterization makes the scale explicit: a linear fit $\log(\mathrm{scale}) = a \log\|x_c\| + b$ exponentiates to $\mathrm{scale} = e^b \|x_c\|^a$, where $a$ is the norm-dependence exponent and $e^b$ is the scale at $\|x_c\|{=}1$ rather than a separately introduced temperature. This form gives well-conditioned gradients in $a$ regardless of $\|x\|$ (Appendix~\ref{sec:cosine-sae}). It also interpolates cleanly between two familiar cases. When $a{=}0$, $\|x_c\|^a{=}1$ and $s_i = e^b \cos(x_c, w_i) + b_{\mathrm{enc},i}$ is pure cosine at scale $e^b$. When $a{=}1$, $s_i = e^b\, \langle x_c, w_i\rangle + b_{\mathrm{enc},i}$ is inner product up to a global scale, since $w_i$ is unit-norm. After applying a ReLU, this score is passed to BatchTopK \cite{bussmann2024batchtopk}. The decoder remains unchanged: $\hat x = z W_{\mathrm{dec}} + b_{\mathrm{dec}}$, with $\mathcal{L} = \|x - \hat x\|^2$. We initialize $a = 0$ and $b = \log\sqrt{d_{\mathrm{model}}}$.

The body tracks the two variants used in the headline 500M comparison. A third corner of the family, pinned cosine ($a{=}0$, where $a$ is frozen and only $b$ trains), is deferred to App.~\ref{app:pinned-cosine}, alongside the full design-space ablation in App.~\ref{app:ablation-matrix}.

\phantomsection\label{sec:method-global}
\textbf{Global learned $a$.} A single scalar $a \in \mathbb{R}$ shared across features, freed at initialization. Pinning $a{=}0$ (pure cosine) discards all norm information from the encoder; since the decoder reconstructs $x$ at full scale, the system must recover magnitude from the sparse code alone, and in practice this collapses (Appendix~\ref{sec:ablation-matrix}). Freeing $a$ lets the optimizer retain exactly as much magnitude as reconstruction requires.

\phantomsection\label{sec:method-perfeature}
\textbf{Per-feature $a_i$.} Different features may want different norm dependence; a position-encoding feature plausibly needs more magnitude than a semantic one. The direct replacement $a \to a_i$, $b \to b_i$ gives $s_i(x) = e^{b_i}\|x_c\|^{a_i} \cos(x_c, w_i) + b_{\mathrm{enc},i}$, adding $\leq 0.1\%$ parameters. Fully free $a_i$ values are unstable at deep layers (cascading to extreme values at L27; Appendix~\ref{sec:ablation-matrix}). We instead parameterize $a_i = a_{\mathrm{base}} + \delta_i$ as a shared base plus per-feature offset (both initialized to $0$); $b_i$ remains fully free. The shared base anchors the distribution while per-feature deltas allow heterogeneity.

\phantomsection\label{sec:method-recipe}
\textbf{Training recipe.} All experiments use the community SAEBench recipe \cite{karvonen2025saebench}: BatchTopK with $k = 80$, Adam at learning rate $5{\cdot}10^{-5}$, the AuxK auxiliary loss of \citet{gao2024scaling} to revive dead features, unit-norm decoder rows, and geometric-median initialization for $b_{\mathrm{dec}}$. The encoder is initialized as the transpose of the decoder. The recipe is held fixed across the standard baseline and all cosine variants; only the encoder score changes. Appendix~\ref{app:recipe} gives the full hyperparameters and recipe-lineage detail.

%% file: sections/body/04-experiments.tex
\begin{figure}[t]
  \centering
  \includegraphics[width=\columnwidth]{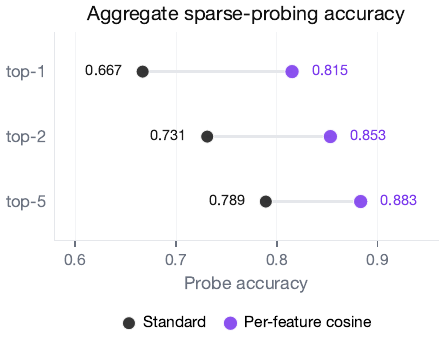}
  \caption{\textbf{Aggregate sparse-probing accuracy.} Top-1, top-2, and top-5 probe accuracy across all eight SAEBench datasets. FVE-matched at $\approx 0.77$. Black: standard SAE; violet: per-feature cosine encoder. The gap narrows at higher $k$ but remains large ($+9.4$\% at top-5). Per-dataset breakdown: Fig.~\ref{fig:hero} Panel A.}
  \label{fig:sparse-probing-short}
\end{figure}

\section{Experiments}
\label{sec:experiments}
\label{sec:results-short}
\label{sec:mechanism-short}

Unless stated otherwise, results use Qwen3-8B at layer~18, 500M FineWeb tokens, $d_{\mathrm{sae}} = 65{,}536$, and the recipe in \S\ref{sec:method-recipe}.
Mechanism sweeps use the same model at smaller budgets (2--50M tokens). Appendix~\ref{sec:efficiency} gives cross-layer and cross-model checks on Qwen L9/L27, Gemma-2-2B, and three LayerNorm models; Appendix~\ref{sec:ablation-matrix} gives the architecture ablation.

\subsection{Reconstruction and sparse probing}
\label{sec:exp-headline}
At the headline scale, reconstruction is tied between the standard BatchTopK SAE and both cosine variants. Aggregate FVE differs by $\leq 0.4$\%, and the SAEBench substitution metrics (KL-score and CE-score, both bounded $[0,1]$ with $1$ meaning the SAE preserves the model's logits exactly) agree to within $0.001$ ($0.984$--$0.985$ KL-score, $0.991$--$0.993$ CE-score; Appendix~\ref{sec:results}). The difference is in feature use.

On the eight SAEBench single-feature top-$k$ probing datasets, averaged over three SAE-training seeds at the full 500M scale, the cosine encoder improves top-1 by $+14.1$\% (global regime) and $+14.6$\% (per-feature regime): top-1 is $0.667 \pm 0.003$ (Standard), $0.808 \pm 0.006$ (global), and $0.813 \pm 0.013$ (per-feature) (Appendix~\ref{app:stats}). The gap dwarfs the seed-to-seed SD and the $\pm 0.63$\% 5-seed probe-training noise; we omit these error terms below and report point estimates.
Single-feature probing directly measures whether individual dictionary elements correspond to human-recognizable concepts \cite{karvonen2025saebench}, the core desideratum of interpretability tools.
The gap is not a capacity artifact: giving the standard encoder $3\times$ more dictionary slots at matched L0 makes dead rates \emph{worse} ($89.4$\% vs.\ $77.4$\%), not better (Figure~\ref{fig:dict-size-control}).
The advantage is also not specific to BatchTopK's batch-wide competition: at matched $L_0$ the cosine encoder improves top-1 over inner-product under per-token TopK ($+5$ to $+7$\%) and AbsTopK ($+12$ to $+14$\%) as well, so it reflects inner-product scoring in general rather than one selector (Appendix~\ref{app:selectors}, Table~\ref{tab:selectors}).
It is likewise robust to the sparsity budget: sweeping $k \in \{10, 160\}$, the top-1 gain holds at $+8$ to $+11$\% with no collapse at low $k$, and per-feature interpretability stays matched throughout (Appendix~\ref{app:k-sweep}, Table~\ref{tab:k-sweep}).
\phantomsection\label{sec:exp-interp}Per-feature interpretability remains matched: at the 500M headline, an LLM-judged describe-then-predict evaluation on $1000$ stratified features per architecture scores $19.2$\% (per-feature cosine), $21.3$\% (global), and $20.1$\% (standard), a $2.1$-point band (Appendix~\ref{app:interp-500m}; per-architecture feature examples in Appendix~\ref{app:qualitative-features}).

The gain is uneven across tasks (Fig.~\ref{fig:sparse-probing-short}). Language and code detection account for the largest gaps, but the per-feature advantage remains $+9.1$\% after removing both. Sentiment is the one reversal: standard is higher by $+3.5$\%, consistent with magnitude carrying task-relevant signal there.
We return to this case in \S\ref{sec:lim-short}.

The learned norm dependence remains far from the inner-product limit (Fig.~\ref{fig:arch-overview}, Table~\ref{tab:headline}). The global $a$ converges to $\approx 0.26$; the per-feature $a_i$ have mean $\approx 0.08$, with no $a_i > 0.5$. Both regimes initialize at $a=0$, so these values are learned rather than imposed. Appendix~\ref{app:stats} reports initialization sweeps and a postnorm-MSE alternative where $a$ becomes negative.

\begin{table}[t]
  \centering
  \small
  \caption{\textbf{Matched reconstruction.} Qwen3-8B L18, 500M tokens. Only the score changes. FVE, probing top-1, and learned $a$ are means over three SAE-training seeds (Appendix~\ref{app:stats}); Q4 FVE is the representative seed, which diagnoses the standard baseline's per-quartile failure (\S\ref{sec:exp-diagnostic}) and is reused for the per-checkpoint analyses in \S\ref{sec:exp-mechanism}. Provenance: Appendix~\ref{app:saebench}.}
  \label{tab:headline}
  \begin{tabular}{lccc}
    \toprule
                    & Standard & Global $a$ & Per-feature \\
    \midrule
    FVE             & 0.770 & 0.769 & \textbf{0.771} \\
    Probing top-1   & $0.667{\scriptstyle\,\pm 0.003}$ & $0.808{\scriptstyle\,\pm 0.006}$ & $\mathbf{0.813}{\scriptstyle\,\pm 0.013}$ \\
    Q4 FVE          & $-184$ & $\mathbf{+0.33}$ & $\mathbf{+0.25}$ \\
    Dead \%         & 0.0   & 0.0   & 0.0 \\
    Learned $a$     & --- & 0.258 & mean $0.076$ \\
    \bottomrule
  \end{tabular}
\end{table}

\subsection{High-norms break inner-product reconstruction}
\label{sec:exp-diagnostic}

Aggregate FVE matches, but the standard SAE fails on the high-norm tail predicted by \S\ref{sec:background}. We bin tokens by $\|x_c\|$ into quartiles and write \emph{Q4} for the highest quartile (the $25\%$ of tokens with the largest $\|x_c\|$). Standard and cosine match on Q1--Q3, but the standard encoder over-activates on Q4: for the standard SAE we trained under the community recipe of \S\ref{sec:method-recipe} (the same SAE used in Table~\ref{tab:headline}), Q4 reconstructions have $9.5{\times}$ the input norm, and Q4 FVE is $-184$.\footnote{The large negative per-quartile number does not contradict aggregate FVE parity: per-quartile FVE normalizes by \emph{within}-quartile variance ($1 - \mathrm{MSE}_q / \mathrm{Var}_q(x)$), and within-Q4 variance is small relative to the global variance that aggregate FVE normalizes against, so Q4's large absolute MSE contributes only modestly to the aggregate numerator.} Both cosine regimes keep Q4 FVE positive (Fig.~\ref{fig:q4-pathology}). An independently trained reference SAE shows the same pathology, with Q4 FVE $= -136$ \citep{karvonen2025saebench}.

This is the failure mode predicted in \S\ref{sec:background}: the inner-product score writes $\|x\|$ into the sparse code, and the decoder turns that scale information into inflated reconstructions, even though downstream RMSNorm has already discarded it.

\subsection{Comparing the learned dictionaries}
\label{sec:exp-mechanism}

The standard SAE does not merely encode content worse; it fails to learn content features at all, spending most of its capacity on features that detect token magnitude instead. The $+14.6$\% sparse-probing gap from \S\ref{sec:exp-headline} persists at matched aggregate FVE, so the two architectures must learn qualitatively different dictionaries. This section identifies which features each one learns.

Dead features do not explain the gap. With the auxiliary loss enabled, both architectures keep $\geq 99\%$ of features alive and match on reconstruction and on LLM-judged per-feature interpretability (within a $2.1$-point band at the 500M headline; Appendix~\ref{app:interp-500m}). The dead-feature margin closes, but the $+14.6$\% sparse-probing gap remains.

\begin{figure}[t]
  \centering
  \includegraphics[width=\columnwidth]{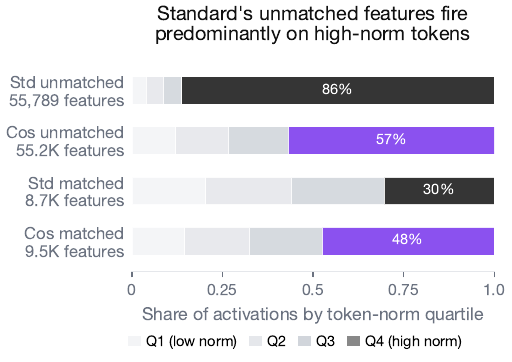}
  \caption{\textbf{Norm-conditioned feature allocation.} Standard's unmatched features fire overwhelmingly on Q4; cosine's unmatched features spread across all quartiles. The standard encoder spends most of its capacity on features whose firing is locked to token norm. Per-quartile reconstruction: Appendix~\ref{app:q4}.}
  \label{fig:norm-conditioned}
  \label{fig:q4-pathology}
\end{figure}

We pair features across the standard and cosine dictionaries when their per-token activation traces have Pearson correlation $\geq 0.7$ \emph{and} their decoder rows have cosine similarity $> 0.7$: i.e., they activate on the same tokens \emph{and} point in the same direction in residual space.
This matches $8{,}661$ features.
The remaining $55{,}789$ standard-only features behave as \emph{norm-conditioned features}: they fire on $\|x\|$ rather than on a content direction.
\begin{itemize}[leftmargin=1.2em,itemsep=1pt,topsep=2pt,parsep=0pt]
  \item \emph{Selective.} $86\%$ of their nonzero activations land in Q4 (vs.\ the $25\%$ a content-blind detector would receive), so they fire on high-norm tokens.
  \item \emph{Strong.} When they do fire on a Q4 token, their mean activation is $48\times$ the per-token mean across the full token population.
  \item \emph{Crowding.} BatchTopK on a high-norm token therefore spends most of its sparse-code budget on these features, leaving little capacity for direction-encoding ones (Fig.~\ref{fig:norm-conditioned}). On tokens where cosine-unique features fire, the standard SAE fires $328$ features/token vs.\ cosine's $122$ (Appendix~\ref{sec:mechanism}).
\end{itemize}

\begin{figure}[t]
  \centering
  \includegraphics[width=\columnwidth]{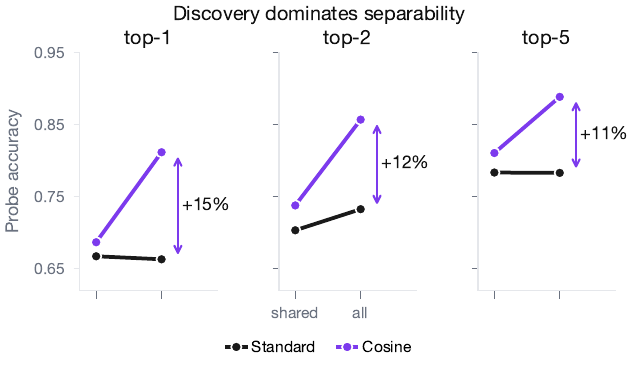}
  \caption{\textbf{Discovery dominates separability.} Sparse-probing accuracy when each SAE uses only features shared with the other dictionary (``shared features'') versus its full dictionary (``all features''). Standard's flat slope shows its unique features add no probe signal; cosine's steep rise shows its unique features encode interpretable concepts. The gap on the right is the total probing advantage, driven almost entirely by feature discovery rather than cleaner encoding of shared directions.}
  \label{fig:discovery-vs-separability}
\end{figure}

Of the standard SAE's $64{,}450$ alive features, only $13.4\%$ direction-encode under this criterion; the rest primarily encode norm. The between-quartile component drives most of the overall cos$>$inner rate ($80$--$87\%$). Within individual quartiles the advantage is a more modest $40$--$70\%$ (Figure~\ref{fig:simpson-paradox}), consistent with norm variation corrupting TopK selection across the full distribution rather than within each quartile independently.
We call these features ``norm-conditioned'' descriptively: they may encode $\|x\|$-correlated information our probing tasks do not measure (auto-interp parity is consistent with this; Appendix~\ref{sec:interpretability}), but they do not encode the eight concept categories sparse probing targets.

\begin{figure}[t]
  \centering
  \includegraphics[width=\columnwidth]{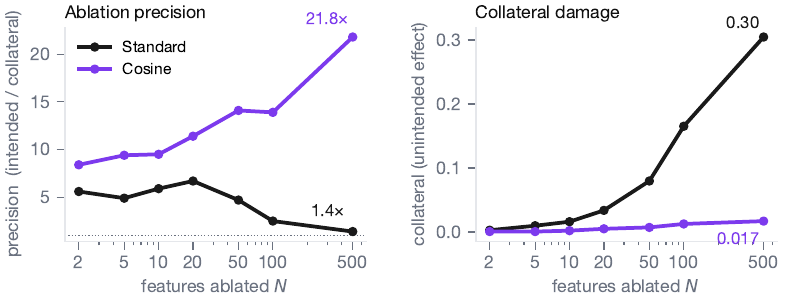}
  \caption{\textbf{Cosine features ablate more cleanly.} Ablating the top-$N$ probe-selected features and decomposing the logit change into intended (target concept) and unintended (collateral) effects. \textbf{Left:} precision (intended/collateral) rises with $N$ for cosine but collapses toward $1\times$ for standard. \textbf{Right:} the cause is collateral, not intended effect; standard's unintended damage grows two orders of magnitude while cosine's stays flat. Qwen3-8B L18, 500M.\DIAG{exp57b}{experiments/57b\_tpp\_collateral/README.md}}
  \label{fig:tpp-precision}
\end{figure}

Restricting both dictionaries to the $8{,}661$ matched features drops the top-1 gap (here the representative seed's $+14.9$\%) to $+2.0$\% (Fig.~\ref{fig:discovery-vs-separability}). Reading $+2.0$\% as the separability contribution and the residual $+12.9$\% as the discovery contribution: $\sim 87\%$ of the gap comes from features the cosine encoder discovers and standard does not, $\sim 13\%$ from cleaner encoding of shared directions. The unmatched standard features add almost nothing to top-5. The unmatched cosine features add $+7.8$\%, confirming that the discovered features encode content, not just norm-correlated firing.
Direct behavioral steering produces matched KL divergence for both architectures (ratio $0.94$--$1.00\times$); the quality gap is which features exist, not their individual intervention power (Appendix~\ref{sec:mechanism}).
Beyond probing, the cosine dictionary also ablates more cleanly (Fig.~\ref{fig:tpp-precision}): removing the top-$N$ probe-selected features, its intended/collateral precision ratio rises with $N$ ($8.4\times \to 21.8\times$) while standard's collapses ($5.6\times \to 1.4\times$). The gap is in collateral damage, not intended effect: both architectures remove the target concept comparably, but standard's unintended side effects grow two orders of magnitude with $N$ (to $0.30$) while cosine's stay flat (to $0.017$), because the norm-conditioned features that fill the standard dictionary are entangled with every concept (full decomposition: Appendix~\ref{sec:mechanism}, Table~\ref{tab:tpp-precision}).

\textbf{Ruling out gradient equalization.}
\label{sec:exp-grad}
A natural alternative is that cosine helps by equalizing gradient magnitudes across norm quartiles (inner-product gradients scale with $\|x_c\|$). The asymmetry is real: the median Q4/Q1 gradient ratio is $1.6$--$2.0\times$ for standard vs.\ $0.8$--$1.0\times$ for cosine. But reweighting the standard loss to equalize gradients by construction closes only $1.9$--$6.8\%$ of the probing gap (Appendix~\ref{app:grad-ratio}); the forward-pass TopK selection geometry, not backward-pass gradient flow, is the lever. A direct score swap confirms this: reading a trained dictionary with the other score (weights fixed) moves probing to match whichever score is used at read time, while reconstruction stays bound to the training score (Appendix~\ref{app:posthoc}, Table~\ref{tab:score-swap}).

Four qualitatively different lines of evidence converge on the same root cause: behavioral (sparse probing), structural (norm-detector characterization), interventional (gradient falsification), and architectural (learned $a$) all point to the inner-product encoder's conflation of direction and magnitude in feature selection. This conflation is a property of the score, not of one selector: it persists under per-token TopK and AbsTopK (Appendix~\ref{app:selectors}), so batch-wide competition amplifies the effect but is not its source.

\ifshownoaux
\subsection{Removing the auxiliary loss for dead features}
\label{sec:aux-k}

The headline runs in \S\ref{sec:exp-headline} keep the auxiliary dead-feature loss \cite{gao2024scaling} enabled in both arms, which equalizes alive-feature counts ($\geq 99\%$) and reconstruction (FVE within $0.4$\%) so that the surviving $+14.6$\% sparse-probing gap cannot be attributed to a dead-feature artifact.
Disabling the auxiliary loss isolates what the score swap does to the dead-feature distribution on its own.
All major public SAE recipes we are aware of \cite{bricken2023monosemanticity,gao2024scaling,karvonen2025saebench} include this loss or an equivalent, so this is a research-only stress test rather than a deployment recipe.

Without the auxiliary loss the cosine advantage survives, but only for the per-feature variant. At Qwen3-8B, 50M tokens, $d_{\mathrm{sae}}{=}16{,}384$ (Fig.~\ref{fig:aux-loss-ablation}), per-feature cosine reduces dead features at every layer (L9 $61.0\to 41.1\%$, L18 $82.8\to 72.4\%$, L27 $78.3\to 28.1\%$) and improves FVE at every layer ($+3.5/+1.6/+7.6$\%). The global-$a$ variant tracks per-feature at L9 and L27 but actually \emph{loses} to Standard at L18 ($85.3\%$ dead vs.\ $82.8\%$): a single learned scale is a poor compromise at L18's mid-range norm, and the per-feature flexibility is what closes that gap. The pattern reproduces on Gemma-2-2B at 50M tokens: the standard encoder leaves $54$--$69\%$ of features dead across L7/L13/L19, while cosine stays at $0.0$--$0.1\%$, with FVE gains of $+3.9$ to $+6.7$\% (Table~\ref{tab:sample-efficiency}; Fig.~\ref{fig:fve-layer}).

\begin{figure}[t]
  \centering
  \includegraphics[width=\columnwidth]{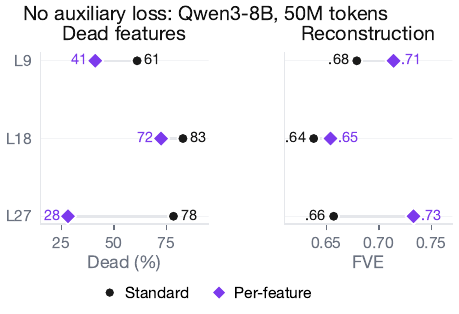}
  \caption{\textbf{Without the auxiliary dead-feature loss, per-feature cosine wins at every layer.} Qwen3-8B, 50M tokens, $d_{\mathrm{sae}}{=}16{,}384$, $\sqrt{d}$ init. Left: dead-feature \%; right: FVE. With the auxiliary loss on (community recipe; headline at $d_{\mathrm{sae}}{=}65{,}536$), both architectures hit $\sim 0\%$ dead and FVE parity (\S\ref{sec:exp-headline}); the gap here is what the auxiliary loss masks. The global-$a$ variant is omitted for readability; see the body.}
  \label{fig:aux-loss-ablation}
\end{figure}

Full convergence curves are in Appendix~\ref{sec:efficiency}.
\fi

\subsection{Cross-model behavior}

\begin{figure}[t]
  \centering
  \includegraphics[width=0.85\columnwidth]{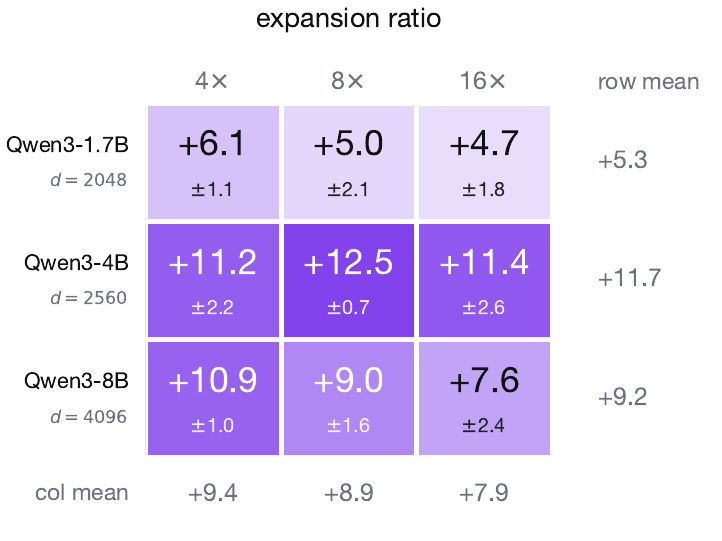}
  \caption{\textbf{Model size, not expansion ratio, drives the cosine advantage.} Sparse-probing top-1 gap (cosine $-$ standard, in percentage points; mean $\pm$ SD over three SAE-training seeds) across the Qwen3 family. Row means (right) show a $\sim 2\times$ jump from 1.7B to $\geq$4B; column means (bottom) are flat. All cells use the community SAEBench recipe, 50M tokens, $k = 80$.\DIAG{exp67}{experiments/67\_scaling\_reseed/README.md}}
  \label{fig:scaling-matrix}
\end{figure}

Our results persist on Gemma-2-2B (a different RMSNorm family, $d_{\mathrm{sae}}=9{,}216$) but do not persist as strongly on non-RMSNorm models.
On Gemma-2-2B, the fixed-SAE top-1 gap is $+3.4{\pm}0.3$\%, smaller than on Qwen but same-signed. To disentangle whether model size or dictionary size drives this, we trained a three-seed $3\times3$ sweep over model size (Qwen3-1.7B/4B/8B) and expansion ratio (4$\times$/8$\times$/16$\times$) at the same recipe (Figure~\ref{fig:scaling-matrix}). The gap tracks model dimension, not dictionary size: row-mean gaps are $+5.3$/$+11.7$/$+9.2$\% across $d_{\mathrm{model}}$, large at every size but not strictly monotonic, while column means across expansion ratio are flat ($+9.4$/$+8.9$/$+7.9$\%). Gemma's smaller gap therefore reflects its small $d_{\mathrm{model}}{=}2304$, not its smaller dictionary.
On three LayerNorm models (Pythia-2.8B/6.9B, Falcon-7B), cosine loses its advantage at deep layers (Appendix~\ref{app:cross-model}). A plausible explanation is that LayerNorm's mean subtraction preserves magnitude-correlated structure that RMSNorm erases: the residual stream's per-coordinate mean carries norm information into the sublayer, so inner-product scoring is partially correct at depth in LayerNorm models. We lack a direct intervention test and flag this as an open question.

The cosine encoder trained on 50M tokens exceeds the standard's 500M sparse-probing top-1 at L9 ($+17.3$\%) and L27 ($+8.7$\%), a $10\times$ token-budget saving (Appendix~\ref{sec:efficiency}).
Norm-conditioned feature allocation and discovery dominance both reproduce across these settings, with the effect magnitude tracking model dimension rather than dictionary size (Appendix~\ref{sec:efficiency}).

%% file: sections/body/05-limitations.tex
\section{Limitations and Future Works}
\label{sec:lim-short}

\emph{Open mechanism gaps and scope bounds.} The mechanism is well isolated but not exhaustively: \S\ref{sec:exp-grad} rules out gradient flow as the load-bearing lever, and a direct score swap in both directions (Appendix~\ref{app:posthoc}) shows reconstruction is training-bound while probing follows the read-time score; an initialization sweep over $a$ remains future work. The decoder is also not norm-invariant: under norm-noise training (Uniform$(0.5, 2.0)$), cosine FVE drops $4.5$--$5.8\%$ vs.\ $1.8$--$3.4\%$ for standard, so a deployment norm shift may erode the gain (Appendix~\ref{sec:limitations}). The gain is also bounded in scope: cosine loses to inner product at deep LayerNorm layers (Pythia-2.8B drops from $100\%$ at L8 to $40\%$ at L24; Appendix~\ref{app:cross-model}), sentiment is the one task reversal (standard $+3.5\%$ top-1), and the 500M headline, though reproduced across three SAE-training seeds (Appendix~\ref{app:stats}), is dictionary-level rather than feature-paired \cite{leask2025canonical}. The advantage holds across three selectors (BatchTopK, per-token TopK, AbsTopK; Appendix~\ref{app:selectors}); penalty-trained selectors (JumpReLU, gated), scale ($>8$B), instruction-tuned/RLHF checkpoints, circuit-level interventions, and competing geometries (QK-norm, nGPT) are deferred to Appendix~\ref{sec:limitations}.

\emph{Future works.} Several recipe choices deserve a closer look. The first is decoder unit-norm: it is load-bearing for reconstruction in some variants (Table~\ref{tab:ablation}), but the constraint is not mirrored in the model's normalized readout, and exploring softer regularizers (norm penalties, scheduled annealing, or absorbing $\|w_i\|$ into a per-feature bias) would tell us whether unit-norm is truly essential or just convenient. A second direction is to fold the per-coordinate RMSNorm gain $g \in \mathbb{R}^d$ into the score itself: the current cosine score $\cos(x_c, w_i)$ ignores $g$, but downstream sublayers read $g \odot x_c$ (\S\ref{sec:background}), so a variant like $\cos(g \odot x_c, w_i)$ would align the encoder strictly with what the model actually consumes. \ifshownoaux Finally, the auxiliary dead-feature loss may be unnecessary under cosine: even without it, cosine retains $+6.3\%$ FVE and $2.39\times$ more alive features than standard at L27 (Appendix~\ref{sec:efficiency}, Table~\ref{tab:no-auxk-summary}), so a cleaner, auxiliary-free recipe looks within reach.\fi

%% file: sections/body/06-discussion.tex
\section{Discussion and Conclusion}
\label{sec:disc-short}

For SAEs trained on normalized residual streams, raw inner-product scoring is the wrong default; the mismatch is starkest under RMSNorm, which erases global scale entirely.
It mixes content alignment with token norm, while the downstream transformer path removes most of that global scale before reading the residual stream.
Replacing the score with a learned norm-scaled cosine keeps reconstruction matched, but changes which features the dictionary spends capacity on.

At matched FVE, KL/CE substitution scores, and per-feature LLM-judged interpretability, our cosine SAE recovers substantially better single-feature coverage of probing targets (\S\ref{sec:exp-mechanism}, \S\ref{sec:exp-interp}).

The evidence points to the forward-pass TopK selection geometry as the primary mechanism.
Equalizing gradient magnitudes closes only $1.9$--$6.8$\% of the gap (\S\ref{sec:exp-grad}): the inner-product score still inflates all pre-activations on high-norm tokens, BatchTopK still selects the same norm-conditioned features, and the same firing pattern reappears.
This mechanism matters because pre-sublayer normalization has already discarded or attenuated token-level magnitude from the downstream computation; the norm-conditioned features that dominate the standard dictionary encode information the model does not fully read.
Whether the gain is best understood as TopK selection geometry alone or as a combination of selection geometry and RMS-geometric alignment remains open; both point to the same practical recommendation: practitioners training SAEs on normalized models should use cosine scoring as the default encoder, with the strongest gains expected under RMSNorm.

We recommend the per-feature base+delta variant ($a_i = a_{\mathrm{base}} + \delta_i$) as the default recipe: it gives the best probing gain ($+14.6$\%), the base+delta anchoring keeps it stable where free per-feature exponents collapse at deep layers (Appendix~\ref{app:perfeature-instability}), and its $1 + 2\,d_{\mathrm{sae}}$ extra parameters are $<0.1\%$ overhead. This is the recipe used for the headline runs; full training details and provenance are in Appendix~\ref{app:recipe}.

Concretely: unit-normalize encoder rows on each forward pass, replace the score with $e^b\|x_c\|^a\cos(x_c, w_i) + b_{\mathrm{enc},i}$, and initialize $a{=}0$, $b{=}\log\sqrt{d_{\mathrm{model}}}$. The global variant adds two scalar parameters; the per-feature variant parameterizes $a_i = a_{\mathrm{base}} + \delta_i$ with per-feature $b_i$, adding $1 + 2\,d_{\mathrm{sae}}$ parameters ($<0.1\%$ overhead).

%% file: sections/03-related-work.tex
\section{Extended Related Work}
\label{sec:related-extended}

This appendix supplements the body Related Work (\S\ref{sec:related}).

\textbf{Residual-stream geometry.} Whitening preprocessing \cite{saraswatula2025whitening} and dimensional collapse in attention outputs \cite{wang2025dimensional}; representation engineering \cite{zou2023repengineering}, sentiment-direction work \cite{tigges2023sentiment}, and invariant subspaces forced by linear interfaces \cite{saurez2026linear}.

\textbf{Adjacent objective-level work.} Latent-scaling crosscoders \cite{minder2025crosscoders} and MonoLoss (vision SAEs) \cite{nasiri2026monoloss}.

\textbf{Alternative SAE benchmarks.} CE-Bench \cite{gulko2025cebench}; SynthSAEBench \cite{chanin2026synthsaebench}.

\textbf{Concurrent failure modes.} Feature absorption \cite{chanin2025absorption} and proposed fixes via masked regularization \cite{narayanaswamy2026masked}, orthogonality \cite{miller2026orthogonality}, hierarchical decoders \cite{luo2026hierarchical}, weight regularization \cite{jedryszek2026stable}; polynomial decoders \cite{koromilas2026polysae}, correlation-shaped superposition \cite{prieto2026correlations}, provable feature extraction \cite{liu2025provably}; sensitivity gaps \cite{tian2025sensitivity}, superficial reasoning features \cite{ma2026falsifying}, pruning robustness \cite{borobia2026pruning}; multi-dimensional features \cite{engels2025multidim}, region-based decompositions \cite{shafran2026directions}.

\textbf{SAE limitations and baselines.} Prompting outperforms SAE steering \cite{wu2025axbench}; raw MLP neurons match SAE sparsity for circuit discovery \cite{arora2026sparse}; LLM baselines match SAE-based model diffing \cite{kempf2026baselines}; SAE features fail to translate into actionable corrections \cite{basu2026actionability}; interpretability illusions OOD \cite{friedman2024illusions}. SAEs remain in use for polysemantic neurons \cite{cunningham2023sparse,grindrod2026holism}; reasoning-process SAEs \cite{yang2026reasoning}.

\textbf{Position relative to prior SAE variants.} Only the encoder score changes; the sparsity mechanism, decoder, and training objective are held fixed. The intervention is compatible with concurrent improvements (Matryoshka, JumpReLU, AbsTopK, masked regularization). Released SAE suites for the Qwen family \cite{qwen2026qwenscope} are available for large Qwen-family runs.

%% file: sections/04-cosine-sae-architecture.tex
\section{Cosine SAE Architecture}
\label{sec:cosine-sae}

This appendix provides the formal variant definitions for reproducibility (\S\ref{app:variant-definitions}), documents the design-space exploration that led to the final architecture (\S\ref{app:design-space}), and reports what the optimizer learns when given freedom (\S\ref{app:empirical-convergence}).


\subsection{Variant Definitions}
\label{app:variant-definitions}
\label{app:architecture-math}

Notation: $x \in \mathbb{R}^d$ residual-stream activation; $x_c = x - b_{\mathrm{dec}}$; $W_{\mathrm{enc}}, W_{\mathrm{dec}} \in \mathbb{R}^{d_{\mathrm{sae}} \times d}$; $\operatorname{normalize}(\cdot)$ is $\ell_2$-normalization along the last axis. All variants share BatchTopK ($k = 80$), optimizer, training data, token budget, and the recipe of \S\ref{app:recipe}.

\paragraph{Standard BatchTopK SAE (baseline).}
\begin{align*}
  z &= \operatorname{BatchTopK}\!\big(\operatorname{ReLU}(x_c W_{\mathrm{enc}}^\top + b_{\mathrm{enc}})\big), \\
  \hat{x} &= z W_{\mathrm{dec}} + b_{\mathrm{dec}}, \qquad
  \mathcal{L} = \|x - \hat{x}\|^2.
\end{align*}
Encoder rows unconstrained: $s_i(x) = \|W_{\mathrm{enc}, i}\|\,\|x_c\|\,\cos(W_{\mathrm{enc}, i}, x_c)$.

\paragraph{Adaptive Cosine SAE (global $a$).}
\begin{align*}
  x_{\mathrm{unit}} &= \operatorname{normalize}(x_c), \\
  W_{\mathrm{unit}} &= \operatorname{normalize}(W_{\mathrm{enc}}, \mathrm{dim}{=}{-}1), \\
  \cos(x_c, w_i) &= (x_{\mathrm{unit}} W_{\mathrm{unit}}^\top)_i, \\
  \gamma(x) &= \exp\!\big(a \log \|x_c\| + b\big), \quad a, b \in \mathbb{R}, \\
  s_i(x) &= \gamma(x)\, \cos(x_c, w_i) + b_{\mathrm{enc}, i}, \\
  z &= \operatorname{BatchTopK}\!\big(\operatorname{ReLU}(s)\big), \\
  \hat{x} &= z W_{\mathrm{dec}} + b_{\mathrm{dec}}, \quad \mathcal{L} = \|x - \hat{x}\|^2.
\end{align*}
At $a = 0$, $s_i = e^b \cos(x_c, w_i)$; at $a = 1$, $s_i = e^b \langle W_{\mathrm{unit}, i}, x_c\rangle$. Init: $a = 0$, $b = \log\sqrt{d_{\mathrm{model}}}$. Added params: $2$ scalars.

\paragraph{Per-Feature Adaptive Cosine SAE (base+delta).}
\begin{align*}
  a_i &= a_{\mathrm{base}} + \delta_i, \quad a_{\mathrm{base}}, \{\delta_i\} \in \mathbb{R}^{1 + d_{\mathrm{sae}}}, \\
  \gamma_i(x) &= \exp\!\big(a_i \log \|x_c\| + b_i\big), \\
  s_i(x) &= \gamma_i(x)\, \cos(x_c, w_i) + b_{\mathrm{enc}, i}.
\end{align*}
Encode/decode/loss as Adaptive Cosine. Init: $a_{\mathrm{base}} = 0$, $\delta_i = 0$, $b_i = \log\sqrt{d_{\mathrm{model}}}$. Added params: $2 d_{\mathrm{sae}} + 1$ scalars ($\leq 0.1\%$). The base+delta parameterization prevents per-feature instability at deep layers (\S\ref{app:perfeature-instability}).

\paragraph{Magnitude-Bypass SAE.}
\begin{align*}
  x_{\mathrm{unit}} &= x_c / \|x_c\|, \quad W_{\mathrm{unit}} = \operatorname{normalize}(W_{\mathrm{enc}}, \mathrm{dim}{=}{-}1), \\
  z &= \operatorname{BatchTopK}\!\big(\operatorname{ReLU}(x_{\mathrm{unit}} W_{\mathrm{unit}}^\top)\big), \\
  W_{\mathrm{dec}}^{u} &= \operatorname{normalize}(W_{\mathrm{dec}}, \mathrm{dim}{=}{-}1), \\
  x_{\mathrm{raw}} &= z W_{\mathrm{dec}}^{u}, \\
  \hat{x} &= \|x_c\| \cdot \frac{x_{\mathrm{raw}}}{\|x_{\mathrm{raw}}\|} + b_{\mathrm{dec}}, \quad \mathcal{L} = \|x - \hat{x}\|^2.
\end{align*}
Encoder activations bounded in $[0, 1]$; magnitude $\|x_c\|$ enters only the decoder rescale. The sparse code carries no magnitude information.

\paragraph{Gradient asymmetry.} For an active feature $i$:
\[
  \frac{\partial s_i^{\mathrm{std}}}{\partial W_{\mathrm{enc}, i}} = x_c
  \;\;\Rightarrow\;\;
  \left\| \frac{\partial s_i^{\mathrm{std}}}{\partial W_{\mathrm{enc}, i}} \right\| = \|x_c\|.
\]
For all cosine variants, $s_i$ depends on $x_c$ through $x_c / \|x_c\|$ and a scalar, so the encoder-gradient norm is bounded independently of input magnitude.

\paragraph{Held fixed across variants.} BatchTopK $k = 80$; geometric-median init of $b_{\mathrm{dec}}$; decoder columns Kaiming + unit-norm; encoder $= W_{\mathrm{dec}}^\top$ ($0.1 \cdot W_{\mathrm{dec}}^\top$ for Standard); Adam, learning rate, schedule; token budget; $d_{\mathrm{sae}}$. All $\dagger$ and unmarked rows use the AuxK auxiliary loss of \citet{gao2024scaling}; the $\ddagger$ rows (from an early 5M-token sweep) predate its adoption but exp46 confirms aux-k is a no-op for cosine-encoder architectures at that budget (\S\ref{app:norm-restore}).


\subsection{Design-Space Exploration}
\label{app:design-space}
\label{app:ablation-matrix}
\label{sec:ablation-matrix}

The final architecture emerged from a series of failures. Each design choice prevents a specific collapse mode. Table~\ref{tab:ablation} summarizes the full ablation; the subsections below explain the rationale.

\begin{table*}[t]
  \centering
  \scriptsize
  \setlength{\tabcolsep}{3.5pt}
  \caption{Design-space ablation. Headline setting: Qwen3-8B L18, $500$M tokens; $\dagger$ from L27/$50$M (same recipe); $\ddagger$ from L27/$5$M (early sweep; no aux-k, but exp46 confirms aux-k is a no-op at this budget). \emph{Cos$>$inner} is the per-feature win rate (\%) on the diagnostic of \S\ref{sec:diagnostic}.\DIAG{exp40/42c/44}{experiments/40\_saprmarks\_recipe/README.md}\DIAG{exp46}{experiments/46\_normscope\_ablation/README.md}\DIAG{exp48}{experiments/48\_input\_norm\_verification/README.md}}
  \label{tab:ablation}
  \begin{tabular}{l|cccccc|ccccc}
    \toprule
    \textbf{Variant} & \multicolumn{6}{c|}{\textbf{Design knobs}} & \multicolumn{5}{c}{\textbf{Metrics}} \\
                & Input norm & Unit enc. & Unit dec. & Restore $\|x\|$ & Scale & Enc.\ bias & FVE & Dead\% & Probe top-1 & Cos$>$inner & Learned $a$ \\
    \midrule
    inner-product baseline           & ---  & ---  & \checkmark & ---  & none  & \checkmark & $0.770$ & $0.0$ & $0.667$ & $62$ & --- \\
    naive cosine ($\times$)          & \checkmark & \checkmark & \checkmark & ---  & none  & \checkmark & $0.009^\ddagger$ & $97.0^\ddagger$ & --- & --- & --- \\
    global $b$, no $a$               & \checkmark & \checkmark & \checkmark & ---  & global $b$ & \checkmark & $0.449^\ddagger$ & $76^\ddagger$ & --- & --- & --- \\
    \textbf{global $a$ encoder}      & \checkmark & \checkmark & \checkmark & ---  & global $a, b$ & \checkmark & $0.769$ & $0.0$ & $0.800$ & $63$ & $0.258$ \\
    per-feature $a_i$ (no anchor)    & \checkmark & \checkmark & \checkmark & ---  & per-feat $a_i, b_i$ & \checkmark & $0.721^\dagger$ & $83.4^\dagger$ & --- & $71^\dagger$ & $0.022$ \\
    \textbf{per-feature base+delta}  & \checkmark & \checkmark & \checkmark & ---  & base$+\delta_i$ & \checkmark & $\mathbf{0.771}$ & $0.0$ & $\mathbf{0.815}$ & $62$ & $0.076$ \\
    no enc.\ norm, per-feat $b_i$    & \checkmark & ---  & \checkmark & ---  & per-feat $b_i$ & \checkmark & $0.770^\dagger$ & $0.4^\dagger$ & --- & $61^\dagger$ & $0.201^\dagger$ \\
    no enc.\ norm, global $a, b$ ($\times$) & \checkmark & ---  & \checkmark & ---  & global $a, b$ & \checkmark & $0.614^\dagger$ & $93.2^\dagger$ & --- & --- & $0.620^\dagger$ \\
    magnitude-bypass                 & \checkmark & \checkmark & \checkmark & \checkmark & none & --- & $0.767$ & $4.3$ & $0.783$ & $69$ & --- \\
    \quad encoder rows free          & \checkmark & ---  & \checkmark & \checkmark & none & --- & $0.754^\dagger$ & $0.3^\dagger$ & --- & $67^\dagger$ & --- \\
    \quad decoder rows free          & \checkmark & \checkmark & ---  & \checkmark & none & --- & $0.550^\ddagger$ & $0.0^\ddagger$ & --- & --- & --- \\
    \quad both encoder/decoder free  & \checkmark & ---  & ---  & \checkmark & none & --- & $0.554^\ddagger$ & $0.0^\ddagger$ & --- & --- & --- \\
    \quad w/o norm restore ($\times$) & \checkmark & \checkmark & ---  & ---  & none & --- & $0.082^\ddagger$ & $88.5^\ddagger$ & --- & --- & --- \\
    input norm only, no scale ($\times$) & \checkmark & ---  & \checkmark & ---  & none  & \checkmark & $0.297^\dagger$ & $93.5^\dagger$ & --- & $58^\dagger$ & --- \\
    no enc.\ norm, base$+\delta_i$   & \checkmark & ---  & ---  & \checkmark & base$+\delta_i$ & \checkmark & $0.740^\dagger$ & $20.0^\dagger$ & --- & $61^\dagger$ & $-0.59^\dagger$ \\
    \bottomrule
  \end{tabular}
\end{table*}

\subsubsection{Why the scale parameter is necessary}
\label{app:why-scale}

The most natural first attempt is pure cosine scoring: normalize both the input and the encoder rows, so $s_i = \cos(x_c, w_i) + b_{\mathrm{enc},i}$. This removes all norm dependence from feature selection. It also fails catastrophically: FVE $= 0.009$ with $97$\% dead features (Table~\ref{tab:ablation}, row ``naive cosine'').\DIAG{exp10}{experiments/10\_cosine\_sae\_training/README.md}

The failure is a reconstruction-scale mismatch. Cosine scores are bounded in $[-1, 1]$; after BatchTopK and ReLU, the sparse code $z$ has entries in $[0, 1]$. The decoder must reconstruct $x$ at its full scale ($\|x_c\| \sim 64$--$400$ depending on layer), but $z W_{\mathrm{dec}}$ produces vectors of order $\|z\|_1 \cdot 1 \approx k = 80$. At shallow layers where $\|x_c\| \approx 64$ this nearly works; at L27 where $\|x_c\| \approx 407$ the decoder cannot bridge the gap. Features receive near-zero gradient because reconstruction error is dominated by the global scale mismatch rather than directional errors, and $97$\% die within the first few thousand steps.

Adding a learned global intercept $b$ (so the scale is $e^b$, a single scalar) partially recovers: FVE rises to $0.449$ but $76$\% of features still die (row ``global $b$, no $a$''). The intercept provides a fixed boost but cannot adapt to the token-level norm variation that the decoder needs. Freeing the exponent $a$ (so scale $= e^{a \log\|x_c\| + b} = e^b \|x_c\|^a$) lets the encoder pass exactly as much per-token magnitude as reconstruction requires. The optimizer converges to $a \approx 0.26$ globally and $\bar{a}_i \approx 0.08$ per-feature; $0$\% dead, FVE $= 0.77$.\DIAG{exp12}{experiments/12\_adaptive\_scaling/README.md}\DIAG{exp42c}{experiments/42c\_noc\_500m/README.md}

\subsubsection{Per-feature instability and the base+delta fix}
\label{app:perfeature-instability}

Allowing each feature its own exponent $a_i$ is appealing: a positional feature may need more norm information than a semantic one. The direct parameterization ($a_i$ free, initialized at $0$) works at shallow layers and short budgets (L9/5M: $0$\% dead, matching the global variant). At L27 with $50$M tokens, it collapses: $83.4$\% dead ($54{,}657$ of $65{,}536$ features). This cascade occurs even with encoder unit-normalization active and is specific to the per-feature degree of freedom in the exponent, not to missing normalization constraints.\DIAG{exp47}{experiments/47\_perfeature\_robustness/README.md}

The failure mechanism is a winner-take-all cascade. At L27, $\|x_c\| \approx 407$ while the initial scale $e^b = \sqrt{d} = 64$, a $6.4\times$ mismatch. Some features randomly receive slightly more gradient early in training (due to initialization variance in $W_{\mathrm{enc}}$ directions), fire more often, and grow their $a_i$ to capture more of the high-norm tokens. As their $a_i$ increases, their pre-activations on high-norm tokens inflate further, raising the batch-wide TopK threshold. Features that started with slightly less gradient now fall below the threshold, receive zero gradient, and die. The cascade is self-reinforcing: at step $5{,}500$, approximately $43{,}000$ features die simultaneously in a single $500$-step window. By step $10{,}000$ the dictionary has $83$\% dead features and cannot recover.

The base+delta parameterization $a_i = a_{\mathrm{base}} + \delta_i$ prevents this cascade. The shared $a_{\mathrm{base}}$ moves all features together during early training, ensuring the global scale tracks $\|x_c\|$ before any per-feature divergence can occur. Once $a_{\mathrm{base}}$ has converged (typically by step $5{,}000$), the per-feature $\delta_i$ offsets begin differentiating. The result: $0.4$\% dead at L27/$50$M (vs.\ $83.4$\% unconstrained), with FVE $= 0.772$ vs.\ $0.721$.\DIAG{exp47}{experiments/47\_perfeature\_robustness/README.md}

\begin{figure}[h]
  \centering
  \includegraphics[width=\columnwidth]{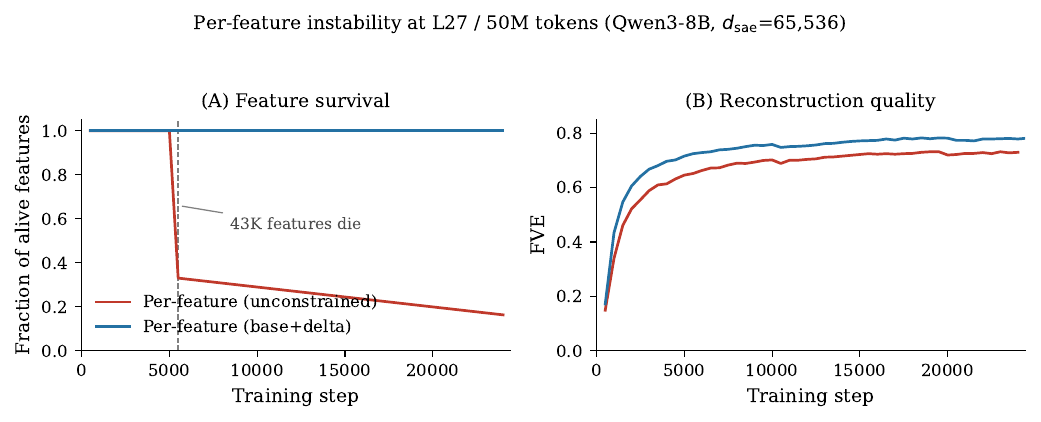}
  \caption{\textbf{Winner-take-all cascade in unconstrained per-feature $a_i$.} Qwen3-8B L27, $50$M tokens. \textbf{(A)} Unconstrained per-feature (red) loses $67$\% of features in a single $500$-step window at step $5{,}500$; base+delta (blue) retains all features throughout. \textbf{(B)} Base+delta also achieves higher final FVE ($0.77$ vs.\ $0.72$) because surviving features encode useful content rather than fighting for norm-dominated TopK slots.\DIAG{exp47}{experiments/47\_perfeature\_robustness/README.md}}
  \label{fig:perfeature-cascade}
\end{figure}

At the headline setting (L18, $500$M tokens), base+delta achieves the best sparse probing in the entire table ($0.815$) at $0$\% dead and matched FVE ($0.771$).

\subsubsection{Stabilization mechanisms: encoder normalization and norm restoration}
\label{app:encoder-norm}
\label{app:norm-restore}

Once input normalization strips $\|x\|$ from the encoder's view, something must prevent norm-dominance from re-entering through other paths. Two mechanisms independently solve this problem: (1) encoder unit-normalization, which prevents encoder-row magnitudes from drifting and recapitulating norm dominance on the weight side; and (2) post-decode norm restoration, which rescales the reconstruction to match $\|x_c\|$ externally. Either one suffices; removing both is catastrophic.

\paragraph{Encoder normalization.} Unit-normalizing encoder rows ($w_i \leftarrow w_i / \|w_i\|$ on each forward pass) ensures the cosine score is a true cosine similarity. Dropping this constraint while keeping input normalization (row ``input norm only, no scale'') gives FVE $= 0.297$ at $93.5$\% dead. The failure mode: without $\|w_i\| = 1$, encoder row norms drift freely. Rows that happen to grow large dominate the TopK competition (their pre-activations scale with $\|w_i\|$), starving smaller rows of gradient. This recapitulates the same norm-dominance pathology we set out to fix, but now on the encoder side rather than the input side. The collapse is layer-independent: free-encoder adaptive cosine dies at $90.8$\% (L9), $94.6$\% (L18), and $93.2$\% (L27), all at $50$M tokens.\DIAG{exp23}{experiments/23\_input\_norm\_ablation/README.md}\DIAG{exp45}{experiments/45\_encoder\_norm\_ablation/README.md}

Releasing encoder normalization while keeping post-decode norm restoration and a global learned $a, b$ (row ``no enc.\ norm, global $a,b$'') also collapses: $93.2$\% dead, with $a$ running away to $0.62$ (toward inner product). The restoration step helps the decoder but cannot prevent the encoder-side norm drift that kills feature selection. The one exception: per-feature $b_i$ with free encoder rows achieves $0.4$\% dead (row ``no enc.\ norm, per-feat $b_i$''). Here each feature's individual scale $e^{b_i}$ can compensate for encoder-row magnitude variation, effectively absorbing $\|w_i\|$ into $b_i$. This works but at a cost: FVE matches the baseline ($0.770$); sparse-probing performance is untested for this variant.\DIAG{exp46}{experiments/46\_normscope\_ablation/README.md}

\begin{figure}[h]
  \centering
  \includegraphics[width=\columnwidth]{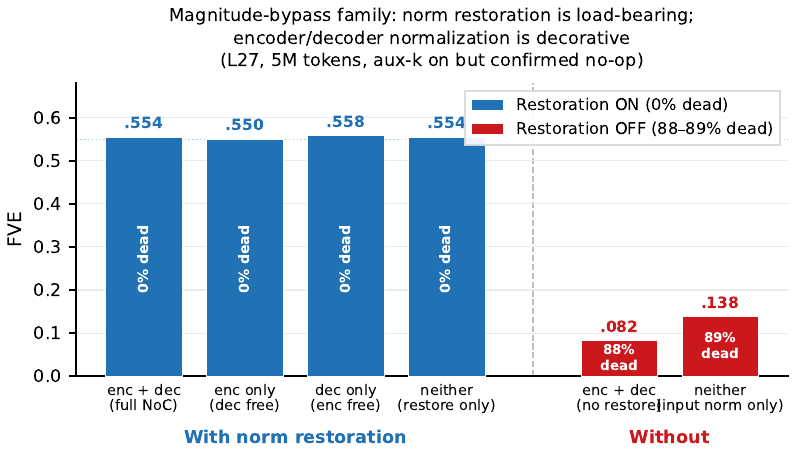}
  \caption{\textbf{Magnitude-bypass family: norm restoration is load-bearing; encoder/decoder normalization is decorative.} L27, $5$M tokens, aux-k on (confirmed no-op). All four restoration-on variants (blue) achieve FVE $\approx 0.55$ at $0$\% dead regardless of enc/dec norm. Restoration-off variants (red) collapse to $\geq 88$\% dead. For the adaptive family, encoder normalization plays the equivalent stabilizing role (see text above).\DIAG{exp46}{experiments/46\_normscope\_ablation/README.md}}
  \label{fig:normalization-factorial}
\end{figure}

\paragraph{Norm restoration.} Within the magnitude-bypass family (no learned $a$, sparse codes bounded in $[0,1]$), norm restoration is the load-bearing stabilizer. A factorial ablation at L27/$5$M shows: restoration-on variants all land within $0.8$\% FVE ($0.550$--$0.558$) at $0$\% dead regardless of encoder/decoder normalization; restoration-off variants collapse to $\sim 89$\% dead with FVE $0.08$--$0.14$ (Figure~\ref{fig:normalization-factorial}). Encoder and decoder unit-norm constraints are decorative when restoration is active.\DIAG{exp46}{experiments/46\_normscope\_ablation/README.md}

\paragraph{Learned exponent as a third path.} With a learned exponent $a$ (the adaptive and per-feature variants), neither encoder normalization nor norm restoration is strictly necessary. The per-feature base+delta variant achieves $0$\% dead and FVE $0.771$ with encoder normalization but without restoration (Table~\ref{tab:ablation}, row 6). Conversely, base+delta without encoder normalization but with restoration achieves FVE $= 0.740$ at $20$\% dead (row ``no enc.\ norm, base+$\delta_i$''); functional but degraded. The learned exponent provides partial self-stabilization: pre-activations scale as $\|x_c\|^{a_i}$, so the sparse code carries per-token magnitude directly and the decoder can reconstruct at the correct scale without an external rescale.

The base+delta parameterization is uniquely robust because the shared $a_{\mathrm{base}}$ anchors all features to a common scale (preventing the encoder-side divergence that kills free-encoder variants) while the per-feature $\delta_i$ offsets allow heterogeneity without competitive pressure. This is why it survives with only one stabilizer active. Other parameterizations (global $a$, unconstrained per-feature) require encoder normalization.

\paragraph{Summary and open questions.} In practice, our recommended variants use encoder normalization (which is cheap and guarantees true cosine geometry). The adaptive/per-feature family does not use norm restoration; the magnitude-bypass family relies on it. We have not ablated all pairwise combinations at the headline setting. Specifically, (1) per-feature base+delta with both encoder normalization and norm restoration active, and (2) per-feature base+delta without either, remain untested at $500$M tokens. Whether the $20$\% dead rate of the no-enc-norm variant (exp48, L27/$50$M) improves with longer training or worsens is unknown. These are natural directions for future exploration.

\subsubsection{Initialization sensitivity}
\label{app:init}

The scale parameter $b$ is initialized to $\log\sqrt{d_{\mathrm{model}}}$, which sets the initial pre-activation magnitude to $\sqrt{d} \cdot \cos(\theta) \approx 64$ for Qwen3-8B. This works when residual-stream norms are of similar order or larger (Qwen L9: $\|x\| \approx 58$, L27: $\approx 407$). It fails catastrophically when norms are much smaller.\DIAG{exp42}{experiments/42\_mistral\_init/README.md}

On Mistral-7B, activation norms at L8 are only $6.3$ ($10\times$ smaller than $\sqrt{d} = 64$). The $\sqrt{d}$ initialization overshoots: pre-activations are an order of magnitude too large, features saturate and die immediately ($100$\% dead at L8, $97.6$\% at L16). No gradient flows from dead features, so the optimizer cannot recover. This is the opposite failure mode from Qwen L27, where $\sqrt{d}$ \textit{undershoots} ($\|x\| = 407$ vs.\ $\sqrt{d} = 64$) but features stay alive and the optimizer climbs toward the correct scale.\DIAG{exp25}{experiments/25\_multimodel\_matrix/README.md}

The asymmetry: undershoot is recoverable (gradients still flow), overshoot is not (dead features produce no gradient). We resolve this with a norm-adaptive initialization rule: if the mean centered activation norm $\operatorname{mean}\|x_{\mathrm{train}} - b_{\mathrm{dec}}\|$ differs from $\sqrt{d_{\mathrm{model}}}$ by more than $2\times$, use $b = \log(\operatorname{mean}\|x\|)$ instead. With this fix, Mistral achieves positive cosine advantage at all layers ($+1.9$\% FVE at L8, $+3.0$\% at L24, $55$\% fewer dead features).\DIAG{exp42}{experiments/42\_mistral\_init/README.md}

A subtlety at longer budgets: on Qwen L27 at $50$M tokens, norm-adaptive init hurts relative to $\sqrt{d}$ because it removes the gradient pressure that teaches $a$ to compensate for scale (the optimizer ``needs to struggle'' with a wrong $b$ to discover that magnitude matters). The $\sqrt{d}$ default works when norms exceed the init; norm-adaptive is necessary only in the overshoot regime.\DIAG{exp34}{experiments/34\_multi\_seed/README.md}\DIAG{exp41}{experiments/41\_optimal\_500m/README.md}

\begin{table}[h]
  \centering
  \caption{Initialization sensitivity across models. The direction of the mismatch determines recoverability.\DIAG{exp42}{experiments/42\_mistral\_init/README.md}\DIAG{exp42c}{experiments/42c\_noc\_500m/README.md}}
  \label{tab:init-cross-model}
  \begin{tabular}{lccccc}
    \toprule
    Model & Layer & $\|x\|$ & Init & Tokens & Outcome \\
    \midrule
    Qwen    & L27 & $407$ & $\sqrt{d}$ (undershoot)    & 500M & best \\
    Qwen    & L27 & $407$ & norm-adaptive & 50M  & hurts ($a$ stuck) \\
    Mistral & L8  & $6.3$ & $\sqrt{d}$ (overshoot)    & 5M   & $100\%$ dead \\
    Mistral & all & varied& norm-adaptive & 50M  & cosine wins \\
    \bottomrule
  \end{tabular}
\end{table}

\subsubsection{Group-size experiments}
\label{app:group-size}

Rather than one global $a$ or $65{,}536$ free $a_i$ values, an intermediate option shares each $a$ across a group of $G$ features. We tested $G \in \{1, 4, 16, 64, 256, 9216\}$ at L13 on Gemma-2-2B/$50$M tokens.\DIAG{exp37}{experiments/37\_regularized\_training/README.md}

$G = 4$ ($4$ groups of $2{,}304$ features) achieves the best SAEBench composite in this sweep: KL-score $0.9785$, CE-score $0.9786$, RAVEL $0.619$, outperforming both the fully global setting ($G = 1$: KL $0.9767$, RAVEL $0.618$) and the fully free setting ($G = 9216$: KL $0.9783$, RAVEL $0.609$). Dead-feature rates decrease monotonically with group size ($41$\% at $G = 1$ to $20$\% at $G = 9216$), but downstream task performance is non-monotone.

We note two caveats. First, this experiment uses Gemma-2-2B at $50$M tokens with a smaller dictionary ($d_{\mathrm{sae}} = 9{,}216$); results may not transfer directly to the headline Qwen3-8B/$65{,}536$ setting. Second, differences between adjacent group sizes are small (within $0.002$ on KL-score), so the $G = 4$ optimum is suggestive rather than definitive. The broader conclusion is more robust: some per-feature heterogeneity in norm dependence improves downstream metrics relative to a single global $a$, while fully unconstrained per-feature freedom can hurt.

\begin{figure}[h]
  \centering
  \includegraphics[width=\columnwidth]{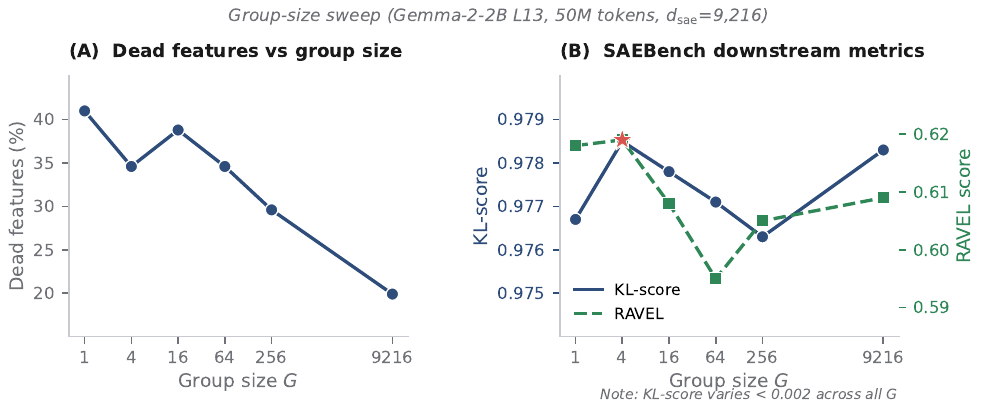}
  \caption{\textbf{Group-size sweep.} Gemma-2-2B L13, $50$M tokens, $d_{\mathrm{sae}} = 9{,}216$. \textbf{(A)} Dead-feature rate decreases monotonically with group size. \textbf{(B)} KL-score and RAVEL peak at $G = 4$ (star), though differences are small ($< 0.002$ KL-score between adjacent sizes). The result is suggestive of an intermediate optimum but has not been replicated at the headline scale.\DIAG{exp37}{experiments/37\_regularized\_training/README.md}}
  \label{fig:group-size-sweep}
\end{figure}

In practice, we use per-feature base+delta rather than group-sharing because: (a) it achieves $0$\% dead at the headline setting; (b) the shared base provides a similar anchoring effect to small group sizes; (c) it adds negligible parameter overhead ($2 d_{\mathrm{sae}}$ scalars). A systematic comparison of group-sharing vs.\ base+delta at the headline scale (Qwen3-8B, $500$M tokens, $d_{\mathrm{sae}} = 65{,}536$) remains a natural future direction.

\subsubsection{Summary of design lessons}
\label{app:design-summary}

\begin{enumerate}[leftmargin=1.5em,itemsep=2pt]
  \item Pure cosine (no scale) fails because the decoder cannot bridge the norm gap between bounded sparse codes and real activation magnitudes (\S\ref{app:why-scale}).
  \item The learned exponent $a$ lets the encoder pass exactly as much magnitude as reconstruction requires; the optimizer consistently chooses $a \ll 1$ (\S\ref{app:empirical-convergence}).
  \item Per-feature $a_i$ must be anchored (base+delta) to prevent winner-take-all cascades at deep layers where $\|x_c\| \gg \sqrt{d}$ (\S\ref{app:perfeature-instability}).
  \item Either encoder unit-normalization or post-decode norm restoration prevents collapse; the magnitude-bypass family requires restoration, while the adaptive family requires encoder norm. Per-feature base+delta can partially survive with only one, but works best with encoder normalization (\S\ref{app:encoder-norm}).
  \item Moderate per-feature freedom (base+delta or $G = 4$) outperforms both fully global and fully free parameterizations (\S\ref{app:group-size}).
  \item Initialization must not overshoot: $\sqrt{d}$ works when norms exceed the init; norm-adaptive is needed otherwise (\S\ref{app:init}).
\end{enumerate}


\subsection{Empirical Convergence}
\label{app:empirical-convergence}

What does the optimizer choose when given freedom? This section reports the learned values of $a$ and $a_i$ at the headline setting.

\begin{table}[h]
  \centering
  \small
  \caption{Learned $a$ across layers (Qwen3-8B, community recipe, 500M tokens).\DIAG{exp42c}{experiments/42c\_noc\_500m/README.md}}
  \label{tab:scale-a}
  \begin{tabular}{llccc}
    \toprule
    Layer & Variant & $a$ (or mean $a_i$) & \% near 0 & Max $a_i$ \\
    \midrule
    L9  & Global          & $0.025$  & ---  & --- \\
    L9  & Per-Feature     & $-0.011$ & --- & $< 0.5$ \\
    L18 & Global          & $0.258$ & --- & --- \\
    L18 & Per-Feature     & $0.076$  & 23\% & $< 0.5$ \\
    L27 & Global          & $0.257$  & ---  & --- \\
    L27 & Per-Feature     & $0.362$  & 23\% & $< 0.5$ \\
    \bottomrule
  \end{tabular}
\end{table}

At L9 the optimizer drives $a$ to near zero (pure cosine); at deeper layers it retains modest norm dependence ($a \approx 0.26$). No feature at any layer or token budget learns $a_i > 0.5$; the inner-product regime ($a = 1$) is never preferred. Per-feature means are consistently lower than the global $a$, suggesting that when features can individually tune their norm dependence, most want less magnitude rather than more.

\begin{figure}[h]
  \centering
  \includegraphics[width=0.85\columnwidth,height=1.7in,keepaspectratio]{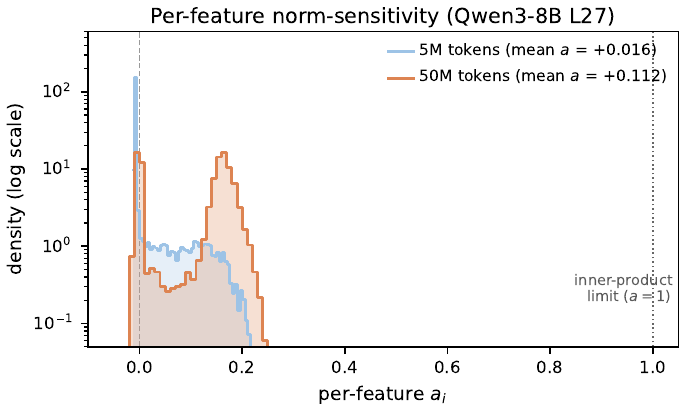}
  \caption{Per-feature $a_i$ distribution at three token budgets (Qwen3-8B L27). Mass shifts toward larger $a_i$ with more data; no $a_i > 0.5$ in any setting.\DIAG{exp42c}{experiments/42c\_noc\_500m/README.md}}
  \label{fig:scale-a-hist}
\end{figure}

\begin{table}[h]
  \centering
  \caption{$\{a_i\}$ statistics across token budgets.\DIAG{exp42c}{experiments/42c\_noc\_500m/README.md}\DIAG{exp16}{experiments/16\_perfeature\_scaling/README.md}\DIAG{exp17}{experiments/17\_production\_scale/README.md}}
  \label{tab:scale-a-summary}
  \begin{tabular}{lcccc}
    \toprule
    Setting & Tokens & Mean $a_i$ & \% near 0 & Max $a_i$ \\
    \midrule
    Qwen L27       & 5M    & $\approx 0.01$ & 88--99\% & $< 0.5$ \\
    Qwen L27       & 50M   & $0.113$ & 60\% & $< 0.5$ \\
    Qwen L27       & 500M  & $0.362$ & 23\% & $< 0.5$ \\
    Qwen L18       & 500M  & $0.076$ & 23\% & $< 0.5$ \\
    \bottomrule
  \end{tabular}
\end{table}

At short budgets ($5$M), $88$--$99$\% of features remain near $a_i \approx 0$. With more data, the distribution shifts: at $500$M only $23$\% stay near zero. This suggests that early in training, features learn content directions first (cosine is sufficient), and only later differentiate their norm dependence as finer-grained reconstruction demands emerge. Even at $500$M, the distribution remains far below the inner-product limit ($a = 1$), confirming that the optimizer consistently prefers direction-dominated scoring.

\phantomsection\label{app:pinned-cosine}
\textbf{Pinned cosine ($a = 0$).} Freezing $a$ at $0$ and training only $b$ produces a fixed-scale cosine encoder. The trained-$\{a_i\}$ distributions above show this is suboptimal: the optimizer wants $a > 0$ at deeper layers, and pinned cosine cannot adapt to per-token norm variation. This variant is included in the ablation table for completeness but is not recommended; global $a$ adds one scalar and strictly dominates.


\subsection{Beyond BatchTopK: Other Selectors}
\label{app:selectors}

The main text uses BatchTopK throughout. Because the cosine score modifies only the encoder pre-activation and not the sparsity mechanism, it composes with any selector. We test whether the cosine advantage is specific to BatchTopK's batch-wide competition or reflects inner-product scoring more generally, by re-running the headline comparison under two additional per-token selectors at the same setting (Qwen3-8B L18, $50$M tokens, $d_{\mathrm{sae}} = 65{,}536$, matched $L_0 = 80$):

\begin{itemize}[leftmargin=1.5em,itemsep=1pt]
  \item \textbf{Per-token TopK} \citep{gao2024scaling}: keep the $k$ largest activations \emph{per token} rather than a single batch-wide budget.
  \item \textbf{AbsTopK} \citep{zhu2025abstopk}: keep the $k$ largest by absolute value per token, preserving sign.
\end{itemize}

\begin{table}[h]
  \centering
  \small
  \caption{\textbf{The cosine advantage persists across selectors, and decomposes into two magnitude channels.} Sparse-probing top-1 at matched $L_0 = 80$ (Qwen3-8B L18, $50$M tokens). The cosine encoder improves top-1 over inner-product under every selector, not only BatchTopK. The \emph{unit-enc} arms are inner-product with $\ell_2$-normalized encoder rows: they remove the weight-norm $\|w_i\|$ but keep the input norm $\|x\|$. Removing $\|w_i\|$ alone rescues features from the dead state (decisively under AbsTopK: $92.4\% \to 0.0\%$) but does \emph{not} recover probing quality, which stays at the inner-product level; only the cosine score, which also removes $\|x\|$, recovers probing.\DIAG{exp63}{experiments/63\_selectors/README.md}}
  \label{tab:selectors}
  \begin{tabular}{llccc}
    \toprule
    Selector & Encoder & FVE & Dead\% & Probe top-1 \\
    \midrule
    BatchTopK      & inner-product     & $0.723$ & $0.0$  & $0.530$ \\
    BatchTopK      & cosine (per-feat) & $0.726$ & $0.0$  & $\mathbf{0.648}$ \\
    \midrule
    per-token TopK & inner-product       & $0.641$ & $91.7$ & $0.731$ \\
    per-token TopK & \;\;+ unit-enc       & $0.680$ & $80.5$ & $0.645$ \\
    per-token TopK & cosine (global)     & $0.711$ & $6.1$  & $\mathbf{0.802}$ \\
    \midrule
    AbsTopK        & inner-product       & $0.636$ & $92.4$ & $0.690$ \\
    AbsTopK        & \;\;+ unit-enc       & $0.670$ & $0.0$  & $0.668$ \\
    AbsTopK        & cosine (per-feat)   & $0.692$ & $6.2$  & $\mathbf{0.827}$ \\
    \bottomrule
  \end{tabular}
\end{table}

\paragraph{The advantage is not BatchTopK-specific.} The cosine encoder improves sparse-probing top-1 over the inner-product encoder under all three selectors (Table~\ref{tab:selectors}): $+11.8$ points under BatchTopK, $+5$ to $+7$ under per-token TopK, and $+12$ to $+14$ under AbsTopK. The effect therefore reflects inner-product scoring in general, not the batch-wide competition specific to BatchTopK.

\paragraph{Batch-wide competition explains part, but not all, of the gap.} Our mechanism account (\S\ref{sec:background}) predicts that per-token selection should \emph{reduce} the cosine advantage: when each token receives exactly $k$ slots, the per-token input-norm scalar multiplies every feature's inner-product score equally and cancels from the within-token ranking, removing the channel by which high-norm tokens claim disproportionate slots. The advantage does shrink under per-token TopK ($+11.8 \to +5$--$7$), consistent with this account. It does not vanish, and under AbsTopK it is as large as under BatchTopK, so a selector-independent component remains. The decomposition below attributes that residual to the second magnitude channel, the encoder-row norm $\|w_i\|$, which per-token selection does not neutralize.

\paragraph{Two magnitude channels: weight norm drives feature death, input norm drives probing quality.} The inner-product score $\|x\|\,\|w_i\|\cos(x, w_i)$ carries magnitude from two sources: the input norm $\|x\|$ and the encoder-row norm $\|w_i\|$. The \emph{unit-enc} arms isolate them, normalizing $w_i$ (removing $\|w_i\|$) while leaving $\|x\|$ in the score; under per-token selection $\|x\|$ multiplies every feature equally and cancels from the within-token ranking, so unit-enc effectively ranks by $\|w_i\|$-free content within each token. Two things separate:
\begin{itemize}[leftmargin=1.5em,itemsep=1pt]
  \item \textbf{Dead features are a weight-norm effect.} Under per-token selection the inner-product encoder loses most of its dictionary ($91.7$\% and $92.4$\% dead, despite the same auxiliary dead-feature loss used everywhere): a few large-norm rows win slots on most tokens and starve the rest. Unit-normalizing the encoder rows rescues them, decisively under AbsTopK ($92.4\% \to 0.0\%$ dead) and partially under per-token TopK ($91.7\% \to 80.5\%$).
  \item \textbf{Probing quality is an input-norm effect.} Rescuing the features does \emph{not} make them more interpretable: unit-enc probing stays at the inner-product level ($0.645$ and $0.668$), far below cosine ($0.802$, $0.827$). Only the cosine score, which additionally strips $\|x\|$, recovers probing. As long as the score scales with token magnitude, the learned features remain norm-conditioned regardless of whether they survive.
\end{itemize}
Cosine scoring is the only encoder here that removes \emph{both} magnitude sources, which is why it alone wins on feature survival \emph{and} probing.

\paragraph{Untested: penalty-trained selectors.} We did not obtain matched-sparsity JumpReLU \citep{rajamanoharan2024jumprelu} or Gated \citep{rajamanoharan2024gated} results. Unlike the fixed-$k$ selectors above, these reach a target $L_0$ through a sparsity penalty rather than a hard budget. At our $50$M-token mechanism-sweep budget the learned threshold initializes correctly at $L_0 \approx 80$, but reconstruction pressure drives it back up within a few hundred steps and the straight-through penalty cannot hold it: across penalty strengths spanning more than a $30\times$ range, and at two dictionary sizes ($d_{\mathrm{sae}} = 16{,}384$ and $65{,}536$), the converged $L_0$ stayed in the thousands rather than near $80$. Reaching matched sparsity requires the substantially longer training schedules used in their original work, so a like-for-like comparison is left to future work.

%% file: sections/05-diagnostic-evidence.tex
\section{Diagnostic: Cosine vs.\ Inner Product as a Causal Predictor}
\label{sec:diagnostic}
\label{app:definitions}

Extended version of \S\ref{sec:exp-diagnostic}. Given fixed decoder directions from a trained Standard SAE, does cosine score have higher absolute correlation with a causal projection-ablation effect than inner product?

\textbf{Definitions.} \textit{Projection ablation}: for unit decoder direction $d_f$, $\mathrm{ablate}(x; f) = x - (d_f^\top x)\, d_f$, re-installed via a forward hook at the ablation layer. \textit{KL-divergence ablation effect}: $y_t = \mathrm{KL}\!\big(\,p_{\mathrm{orig}}(\cdot \mid x_t) \,\|\, p_{\mathrm{abl}}(\cdot \mid x_t; f)\,\big)$.

\begin{figure}[h]
  \centering
  \includegraphics[width=0.85\linewidth]{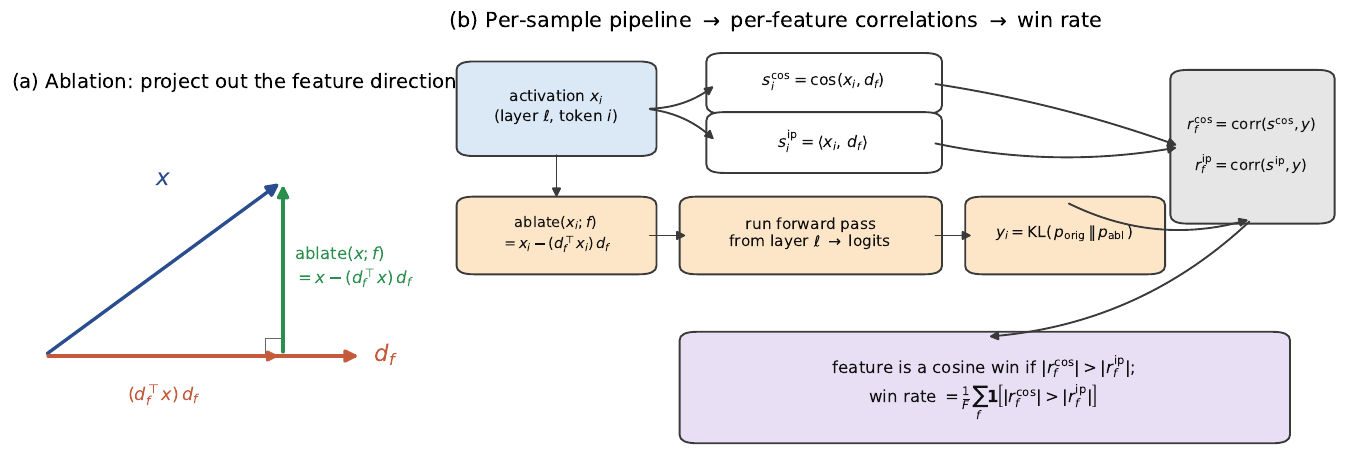}
  \caption{Win-rate diagnostic protocol. \textbf{(a)} Projection ablation removes a decoder direction from the residual stream; the downstream KL-divergence measures its causal importance. \textbf{(b)} Per-sample cosine and inner-product scores are correlated with this causal effect; the scoring rule with higher absolute correlation ``wins'' for that feature.}
  \label{fig:win-rate-explainer-app}
\end{figure}

\textbf{Win-rate protocol.} For each feature $f$ with unit decoder direction $d_f$ and $m$ held-out tokens: $s^{\cos}_t = \cos(x_t, d_f)$, $s^{\mathrm{ip}}_t = \langle x_t, d_f\rangle$, and $y_t$ is the KL-divergence ablation effect defined above. Decoder directions $d_f$ are taken from trained Standard SAEs. Cosine wins iff $|\mathrm{corr}(s^{\cos}, y)| > |\mathrm{corr}(s^{\mathrm{ip}}, y)|$, averaged over features. The $50$\% null corresponds to no systematic advantage for either scoring rule. Projection ablation makes the perturbation magnitude exactly $\langle x, d_f\rangle$, biasing KL toward $|\langle x, d_f\rangle|$, which means the diagnostic is \textit{conservative} with respect to cosine (Figure~\ref{fig:win-rate-explainer-app}).

\begin{figure}[h]
  \centering
  \includegraphics[width=0.95\columnwidth]{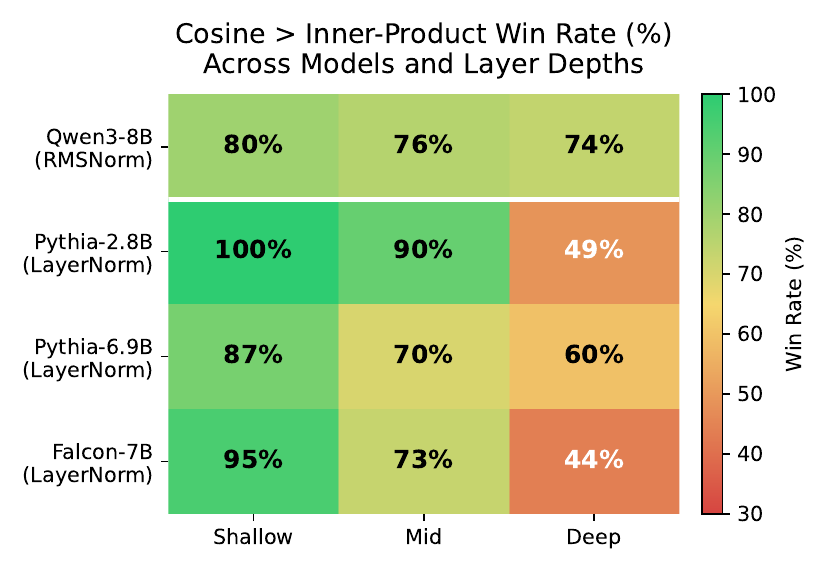}
  \caption{Cosine-vs.-inner-product win rate by depth. Deep RMSNorm: 70--90\%. Deep LayerNorm: at chance.\DIAG{exp25}{experiments/25\_multimodel\_matrix/README.md}}
  \label{fig:cross-model}
\end{figure}

\textbf{Depth dependence on RMSNorm.} Qwen3-8B: L9 $\approx 44\%$, L18 $\approx 65\%$, L27 $\approx 78\%$. Norm-patching at L18 produces $\sim 10\times$ less downstream-KL change than direction-patching at fixed norm. Cross-model RMSNorm: Mistral-7B 74--88\%; Gemma-2-2B matches Qwen with a low value at L20. LayerNorm (Pythia-2.8B / Pythia-6.9B / Falcon-7B) drops from $\sim 90\%$ shallow to 40--57\% deep (Fig.~\ref{fig:cross-model}).\DIAG{exp25}{experiments/25\_multimodel\_matrix/README.md} On constructed token pairs (high-cosine / low-norm vs.\ low-cosine / high-norm), the high-cosine token has the larger KL effect on $\sim 90\%$ of features at deep layers.

\begin{figure}[h]
  \centering
  \includegraphics[width=\columnwidth]{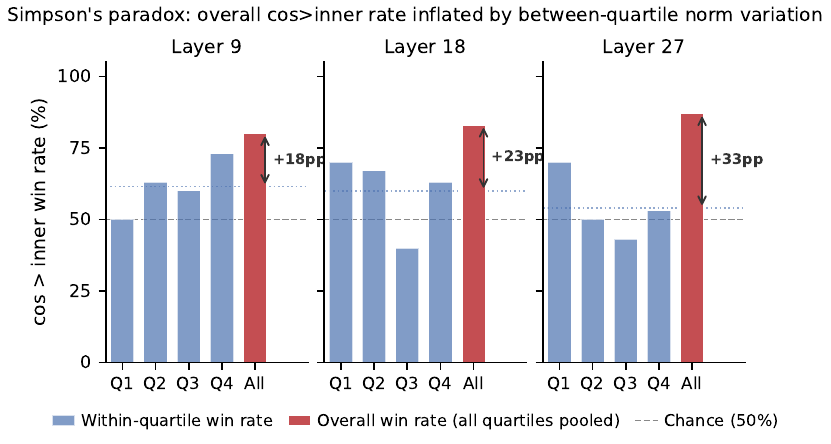}
  \caption{\textbf{Simpson's paradox in the cos$>$inner diagnostic.} Within individual norm quartiles (blue), the win rate hovers at $40$--$70\%$; the overall rate (red) is $80$--$87\%$. The gap grows with depth as residual-stream norm variation increases. This confirms that the between-quartile component (norm variation corrupting cross-token TopK selection) drives most of the overall cos$>$inner rate. Standard SAE, Qwen3-8B, 50M tokens.\DIAG{exp19}{experiments/19\_norm\_stratified/README.md}}
  \label{fig:simpson-paradox}
\end{figure}

\textbf{Caveats.} Within a single norm quartile the rate drops to 40--70\%, so the 70--90\% rate is partly between-quartile (Simpson's paradox; Figure~\ref{fig:simpson-paradox}); the training-time results in \S\ref{sec:exp-headline} do not depend on this measurement. The inflation gap grows with depth ($+18$\% at L9, $+23$\% at L18, $+33$\% at L27), consistent with deeper layers having more norm variation driving the between-quartile effect. The magnitude-bypass 69\% rate at 500M (\S\ref{sec:exp-mechanism}) does not replicate at 5M tokens (27--53\% across four restoration-on variants).\DIAG{exp44}{experiments/44\_norm\_stratified\_fve/README.md}\DIAG{exp46}{experiments/46\_normscope\_ablation/README.md}\DIAG{exp19}{experiments/19\_norm\_stratified/README.md}

\subsection{SAE-Free Direction vs.\ Norm Patching}
\label{app:direction-patching}

\begin{figure}[h]
  \centering
  \includegraphics[width=0.85\columnwidth]{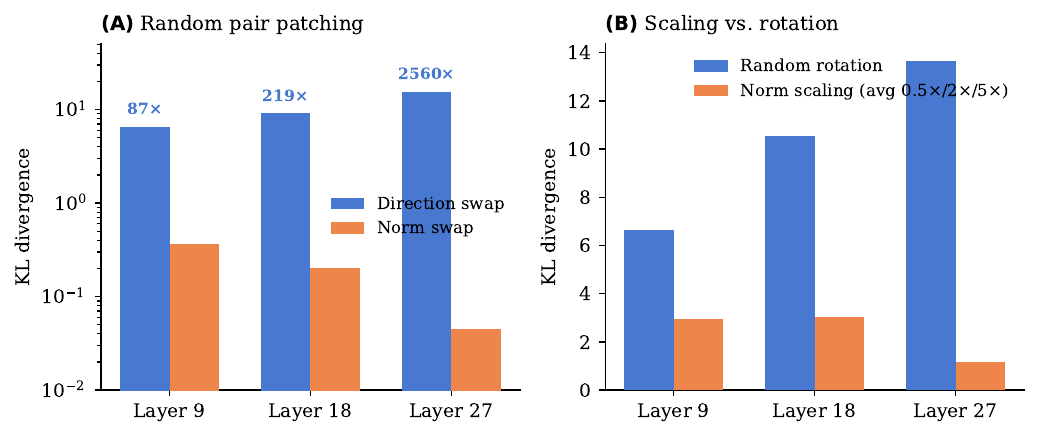}
  \caption{\textbf{The model reads direction, not magnitude.} KL-divergence caused by swapping direction (blue) vs.\ norm (orange) between random token pairs at three Qwen3-8B layers. Direction patches cause $87$--$2{,}560\times$ more disruption; the ratio grows with depth as fewer subsequent layers remain for magnitude to influence direction. $n = 500$ pairs per layer.\DIAG{exp8}{experiments/08\_activation\_patching/README.md}}
  \label{fig:direction-patching}
\end{figure}

The cos$>$inner diagnostic in \S\ref{sec:diagnostic} relies on SAE decoder directions. This subsection removes that dependency entirely: for $500$ random cross-prompt token pairs at each layer, we decompose the residual-stream activation into direction ($\hat x = x/\|x\|$) and magnitude ($\|x\|$), then measure KL-divergence from the unpatched output after swapping each component independently (Figure~\ref{fig:direction-patching}).

\textbf{Scaling test.} Replacing a single token's direction with a random unit vector (at fixed norm) causes $2.2\times$ (L9), $3.5\times$ (L18), and $11.6\times$ (L27) more KL than scaling its norm by $0.5$--$5\times$ at fixed direction. The depth gradient is consistent with RMSNorm progressively erasing magnitude from the sublayer-input path.

%% file: sections/06-production-results.tex
\section{Sparse-Probing Coverage at Matched FVE}
\label{sec:results}

\textbf{Setup.} \texttt{Qwen/Qwen3-8B-Base} layer 18, 500M FineWeb tokens, $d_{\mathrm{sae}} = 65{,}536$, recipe in \S\ref{app:recipe}. Four architectures: Standard, Adaptive Cosine SAE, Per-Feature Adaptive Cosine SAE, Magnitude-Bypass SAE. Body presentation: \S\ref{sec:exp-headline}.

\textbf{Not held fixed across variants.} Encoder score range (cosine $\in [-1, 1]$ vs.\ unbounded inner product); learned scale parameter $a$ ($\leq 0.1\%$ extra parameters); encoder initialization (norm-aware vs.\ $\sqrt{d}$); activation dynamic range; wall-clock time (8--14\% overhead).

\begin{table}[t]
  \centering
  \small
  \caption{Core SAEBench metrics, Qwen3-8B L18, 500M tokens, $d_{\mathrm{sae}} = 65{,}536$.\DIAG{exp40/42c/44}{experiments/40\_saprmarks\_recipe/README.md}}
  \label{tab:saebench-core}
  \begin{tabular}{lcccc}
    \toprule
                    & Std.\  & Adapt.\ Cos. & PF Adapt.\ Cos. & Mag.-Byp.\ \\
    \midrule
    FVE             & $0.770$ & $0.769$ & $\mathbf{0.771}$ & $0.767$ \\
    KL-div score    & $0.985$ & $0.985$ & $0.985$ & $0.984$ \\
    CE-loss score   & $0.993$ & $0.993$ & $0.993$ & $0.991$ \\
    Dead \%         & $0.0$   & $0.0$   & $0.0$   & $4.3$   \\
    \midrule
    Probing top-1   & $0.667$ & $0.800$ & $\mathbf{0.815}$ & $0.783$ \\
    Probing top-2   & $0.731$ & $\mathbf{0.853}$ & $0.853$ & $0.798$ \\
    Probing top-5   & $0.789$ & $\mathbf{0.891}$ & $0.883$ & $0.805$ \\
    Overall         & $0.944$ & $0.957$ & $\mathbf{0.959}$ & $0.914$ \\
    \bottomrule
  \end{tabular}
\end{table}

\textbf{Reconstruction parity.} FVE $\in [0.767, 0.771]$; KL-div score $\in [0.984, 0.985]$; CE-loss score $\in [0.991, 0.993]$; dead features $\leq 4.3\%$.

\textbf{Sparse probing.} Top-1 differences from Standard (mean over three SAE-training seeds): $+14.1\%$ (Adaptive Cosine SAE) and $+14.6\%$ (Per-Feature Adaptive Cosine SAE); Magnitude-Bypass SAE is $+11.6\%$ (single seed). Per-dataset breakdown: Table~\ref{tab:per-dataset-probing}. The dataset-removal robustness table below is computed on the single representative seed ($+13.3\%$ / $+14.9\%$):

\begin{table}[h]
  \centering
  \small
  \caption{Top-1 vs.\ Standard, with high-margin datasets removed.\DIAG{exp40}{experiments/40\_saprmarks\_recipe/README.md}}
  \label{tab:probing-robustness}
  \begin{tabular}{lccc}
    \toprule
    Removed & Adapt.\ Cos. & PF Adapt.\ Cos. & Mag.-Byp. \\
    \midrule
    none                          & $+13.3\%$ & $+14.9\%$ & $+11.6\%$ \\
    europarl                      & $+10.3\%$ & $+12.0\%$ & $+8.6\%$ \\
    europarl + github-code        & $+9.0\%$  & $+9.1\%$  & $+5.6\%$ \\
    \bottomrule
  \end{tabular}
\end{table}

\textbf{Decoder direction overlap.} Mutual nearest-neighbor matching: 73--78\%. Strict $> 0.95$-cosine $+$ Jaccard: 6--17\%. Restricted to alive-and-paired subset at $n = 3$--$6$: $< 1\%$ (\S\ref{sec:limitations}). Headline comparisons are not paired feature-to-feature tests.\DIAG{exp56b}{experiments/56b\_feature\_overlap/README.md}

\textbf{Per-feature interpretability.} Matched across architectures at the 500M headline (describe-then-predict, $1{,}000$ features/arm): $19.2\%$ (Per-Feature Adaptive Cosine SAE), $21.3\%$ (Adaptive Cosine SAE), $20.1\%$ (Standard) (\S\ref{sec:interpretability}). \ifshownoaux Without the auxiliary loss, the alive-feature gap reopens and total interpretable features differ by $\approx 4.5\times$ at matched per-feature rate (\S\ref{app:interpretability}).\fi

\textbf{Compute overhead.} Training wall-clock vs.\ Standard: $+8\%$ (Adaptive Cosine SAE), $+10\%$ (Per-Feature Adaptive Cosine SAE), $+14\%$ (Magnitude-Bypass SAE). Inference: one extra $\log\|x_c\|$ + $\exp$ per token before BatchTopK (Adaptive Cosine SAE, Per-Feature Adaptive Cosine SAE); post-decode norm-and-multiply for Magnitude-Bypass SAE.

\subsection{Per-Dataset Sparse Probing}
\label{app:per-dataset}

Table~\ref{tab:per-dataset-probing} reports the single-feature top-1 by dataset for all four architectures. Per-Feature Adaptive Cosine SAE has higher or equal top-1 to Standard on seven of eight datasets; amazon\_sentiment is the one dataset on which Standard is higher.

\begin{table}[h]
  \centering
  \caption{Single-feature top-1 by dataset.\DIAG{exp40}{experiments/40\_saprmarks\_recipe/README.md}}
  \label{tab:per-dataset-probing}
  \begin{tabular}{lcccc}
    \toprule
    Dataset & Std.\ & Adapt.\ Cos. & PF Adapt.\ Cos. & Mag.-Byp.\ \\
    \midrule
    europarl (languages)        & $0.644$ & $0.986$ & $\mathbf{0.992}$ & $0.978$ \\
    github-code (programming)   & $0.515$ & $0.700$ & $\mathbf{0.813}$ & $0.784$ \\
    bias\_in\_bios set 1        & $0.669$ & $0.798$ & $\mathbf{0.862}$ & $0.786$ \\
    bias\_in\_bios set 2        & $0.559$ & $0.762$ & $0.758$ & $\mathbf{0.766}$ \\
    bias\_in\_bios set 3        & $0.612$ & $\mathbf{0.778}$ & $0.766$ & $0.763$ \\
    amazon\_reviews categories  & $0.712$ & $0.713$ & $\mathbf{0.715}$ & $0.675$ \\
    amazon\_sentiment           & $0.915$ & $\mathbf{0.924}$ & $0.880$ & $0.795$ \\
    ag\_news (topics)           & $0.705$ & $\mathbf{0.736}$ & $0.735$ & $0.720$ \\
    \bottomrule
  \end{tabular}
\end{table}

\subsection{Per-Architecture Sparse-Probing Bars}
\label{app:saebench}

\begin{figure}[h]
  \centering
  \includegraphics[width=0.78\columnwidth,height=2.8in,keepaspectratio]{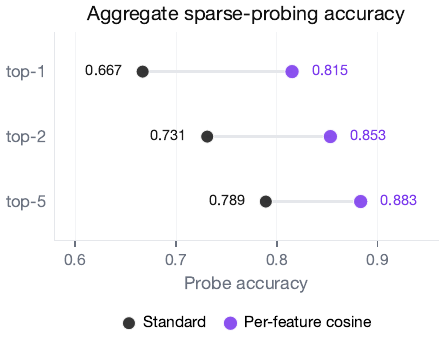}
  \caption{Per-feature top-1 by dataset (Standard vs.\ Per-Feature Adaptive Cosine SAE) and aggregate top-$k$.\DIAG{exp40}{experiments/40\_saprmarks\_recipe/README.md}}
  \label{fig:sparse-probing}
\end{figure}

Figure~\ref{fig:sparse-probing} plots the per-dataset top-1 numbers from Table~\ref{tab:per-dataset-probing} alongside aggregate top-1/2/5 across the eight datasets, comparing Standard against Per-Feature Adaptive Cosine SAE; this is the provenance figure referenced from Fig.~\ref{fig:hero} and Table~\ref{tab:headline}.

%% file: sections/06b-sample-efficiency.tex
\section{Sample Efficiency and Cross-Model Generalization}
\label{sec:efficiency}

\ifshownoaux
\textbf{Without auxiliary loss.} The headline of \S\ref{sec:exp-headline} keeps the auxiliary loss enabled in both arms. With the auxiliary loss removed, the alive-feature differences become visible: on Gemma-2-2B at 50M tokens, Standard shows 54--69\% dead vs.\ $\approx 0\%$ for Adaptive Cosine SAE (Table~\ref{tab:sample-efficiency}). At 500M on Qwen3-8B with $\sqrt{d}$ init, $d_{\mathrm{sae}} = 16{,}384$, no auxiliary loss, the global-$a$ Adaptive Cosine SAE shows $+6.3\%$ FVE and $2.39\times$ alive features over Standard at L27 (Table~\ref{tab:no-auxk-summary}). Both architectures plateau by 300M tokens.

\begin{table}[t]
  \centering
  \small
  \caption{Reference runs at 500M Qwen3-8B. Aux: with the auxiliary loss; No-aux: without.\DIAG{exp40}{experiments/40\_saprmarks\_recipe/README.md}\DIAG{exp42c}{experiments/42c\_noc\_500m/README.md}}
  \label{tab:no-auxk-summary}
  \begin{tabular}{llllp{0.30\linewidth}}
    \toprule
    Run & Layers & $d_{\mathrm{sae}}$ & Recipe & Cosine vs.\ Standard \\
    \midrule
    A & L9, L18, L27 & 16k & No-aux                    & init-contaminated at L18/L27 \\
    B & L18          & 65k & Aux                       & Standard reference, $0\%$ dead \\
    C & L9, L18, L27 & 16k & No-aux, $\sqrt{d}$ init   & $+6.3\%$ FVE, $2.39\times$ alive at L27 \\
    D & L18          & 65k & Aux                       & FVE parity \\
    \bottomrule
  \end{tabular}
\end{table}

\begin{figure}[t]
  \centering
  \includegraphics[width=0.85\columnwidth]{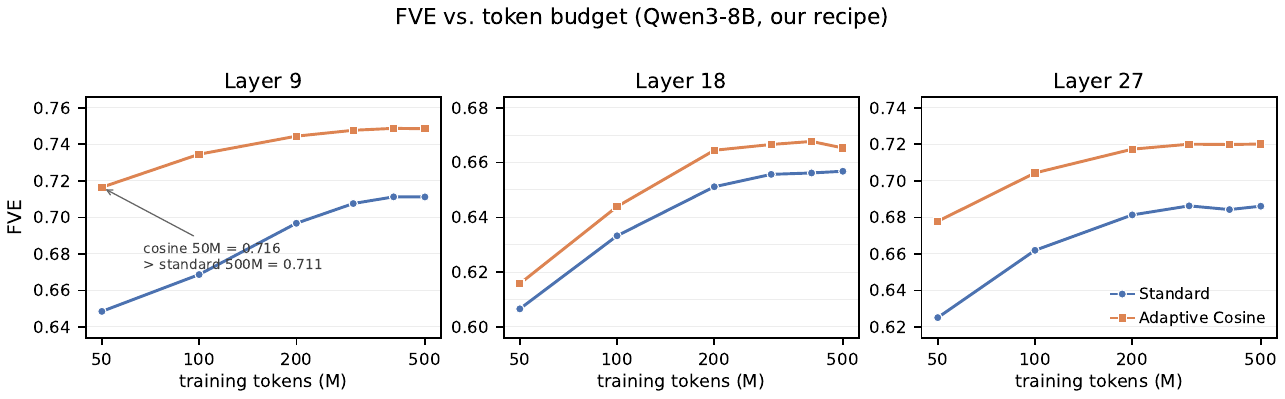}
  \caption{FVE vs.\ token budget on Qwen3-8B (500M, our recipe). Adaptive Cosine SAE at the 50M checkpoint reaches Standard's 500M FVE at L9.\DIAG{exp42c}{experiments/42c\_noc\_500m/README.md}}
  \label{fig:convergence}
\end{figure}

\begin{table}[t]
  \centering
  \small
  \caption{Without auxiliary loss, Gemma-2-2B, 50M tokens.\DIAG{exp35}{experiments/35\_gemma\_saebench/README.md}}
  \label{tab:sample-efficiency}
  \begin{tabular}{llcccc}
    \toprule
    Layer & Variant & FVE & Dead \% & Alive & Gini \\
    \midrule
    L7  & Standard       & $0.759$ & $54.2$ & $4{,}176$  & $0.890$ \\
    L7  & Adapt.\ Cos.   & $\mathbf{0.798}$ & $0.1$  & $\mathbf{9{,}098}$ & $0.662$ \\
    L13 & Standard       & $0.696$ & $68.8$ & $2{,}802$  & $0.906$ \\
    L13 & Adapt.\ Cos.   & $\mathbf{0.747}$ & $0.0$  & $\mathbf{9{,}024}$ & $0.648$ \\
    L19 & Standard       & $0.788$ & $60.2$ & $3{,}616$  & $0.875$ \\
    L19 & Adapt.\ Cos.   & $\mathbf{0.855}$ & $0.0$  & $\mathbf{9{,}161}$ & $0.651$ \\
    \bottomrule
  \end{tabular}
\end{table}
\fi

\textbf{Cross-model coverage.} RMSNorm: Qwen3-8B, Gemma-2-2B, Mistral-7B. LayerNorm: Pythia-2.8B, Pythia-6.9B, Falcon-7B, Pythia-70M. Token budgets vary (500M for the Qwen L18 run, 50M for most cross-model rows, 5M for some); rows in Table~\ref{tab:cross-model-main} are not normalized for budget or recipe.

\begin{table}[t]
  \centering
  \small
  \caption{Cross-model behavior. \emph{cos $>$ inner}: per-feature win rate on the diagnostic of \S\ref{sec:diagnostic} (50\% null); ``$a \to b$'' on LayerNorm rows is shallow$\to$deep. \emph{FVE $\Delta$}: cosine $-$ standard, in absolute FVE units. \emph{Alive $\times$}: alive-feature ratio of cosine over standard.\DIAG{exp25}{experiments/25\_multimodel\_matrix/README.md}\DIAG{exp35}{experiments/35\_gemma\_saebench/README.md}}
  \label{tab:cross-model-main}
  \begin{tabular}{llccc}
    \toprule
    Model & Norm & cos $>$ inner & FVE $\Delta$ & Alive $\times$ \\
    \midrule
    Qwen3-8B    & RMS & 62--89\%               & $+0.6$ to $+8.0$ & 1.7--3.3 \\
    Gemma-2-2B  & RMS & 52--90\%               & $+2.5$ to $+6.7$ & 1.4--3.2 \\
    Mistral-7B  & RMS & 74--88\%               & $+0.6$ to $+3.0$ & varies \\
    Pythia-2.8B & LN  & 100$\to$40             & $+0.8$ to $+5.0$ & $2.3$ (L24) \\
    Pythia-6.9B & LN  & 83--90$\to$50--70      & $+1.8$ to $+3.3$ & varies \\
    Falcon-7B   & LN  & 90--100$\to$40--47     & $+5.0$ (L24)     & $4.5$ (L8) \\
    \bottomrule
  \end{tabular}
\end{table}

\textbf{Variance.} $n = 3$ seeds at Qwen L27, 50M tokens: FVE $0.737$ (Adaptive Cosine SAE) vs.\ $0.657$ (Standard); seed-wise SD $< 0.001$ in both arms; alive-feature ratio $3.26\times$.\DIAG{exp34}{experiments/34\_multi\_seed/README.md} At the 500M L18 headline, three SAE-training seeds give a stable top-1 gap of $+14.1$\% (global) and $+14.6$\% (per-feature), FVE matched to $\pm 0.0002$ (Appendix~\ref{app:stats}).\DIAG{exp61}{experiments/61\_multiseed\_500m/README.md} Cos $>$ inner above 50\% at L27: $p < 2.4 \times 10^{-6}$. Per-feature interpretability rate difference (50M/L27): $p = 0.88$; matched also at the 500M headline (\S\ref{app:interp-500m}).\DIAG{exp33}{experiments/33\_feature\_interpretability/README.md}\DIAG{exp62}{experiments/62\_interp\_causal\_headline/README.md}

\subsection{Depth Dependence Within Qwen3-8B}
\label{app:depth-dependence}

\begin{figure}[h]
  \centering
  \includegraphics[width=\textwidth]{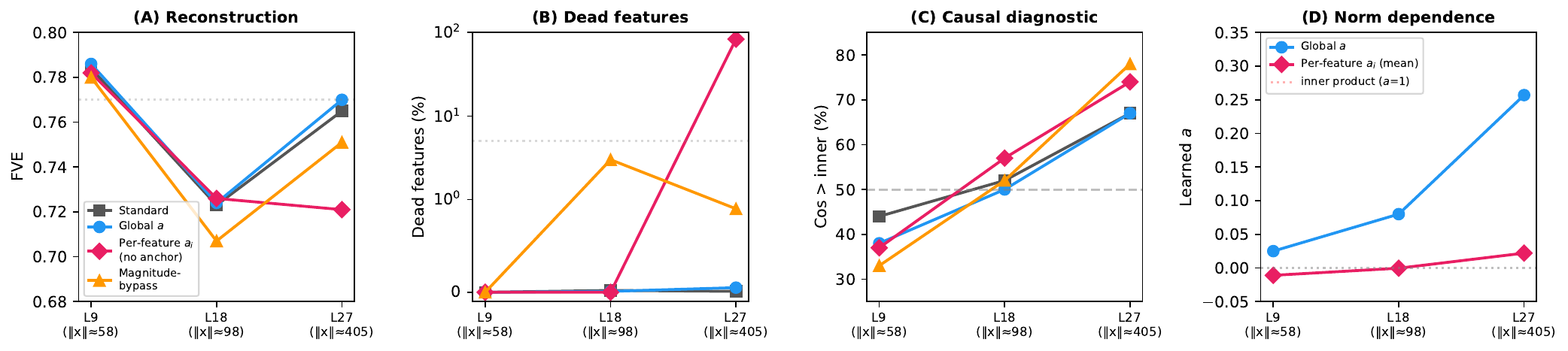}
  \caption{\textbf{Architecture behavior across depth.} Qwen3-8B, $50$M tokens, all four main variants. \textbf{(A)} FVE converges at L9 and diverges at L27. \textbf{(B)} Per-feature (no anchor) collapses catastrophically at L27; magnitude-bypass shows mild dead features at L18. \textbf{(C)} Cos$>$inner crosses $50$\% between L9 and L18, reaching $78$\% for magnitude-bypass at L27. \textbf{(D)} The optimizer drives $a$ toward zero at shallow layers and toward $\sim 0.26$ at deep layers, never approaching $a = 1$.\DIAG{exp43c/43d/43}{experiments/43\_4arch\_50m\_l27/README.md}}
  \label{fig:depth-dependence}
\end{figure}

Figure~\ref{fig:depth-dependence} shows the four main variants at L9, L18, and L27 (all $50$M tokens, same recipe). The key observations:

\begin{enumerate}[leftmargin=1.5em,itemsep=2pt]
  \item \textit{Architectures converge at shallow layers.} At L9 ($\|x\| \approx 58 \approx \sqrt{d}$), all four variants achieve $0$\% dead features and FVE within $0.6$\%. The norm/sqrt($d$) ratio is near unity, so the scale mismatch that drives divergence at deeper layers does not arise.
  \item \textit{Cos$>$inner is below $50$\% at L9.} Inner product is a better causal predictor at L9 ($33$--$44$\%), consistent with magnitude carrying genuine signal at shallow layers where RMSNorm has not yet accumulated. The diagnostic crosses $50$\% between L9 and L18, and reaches $67$--$78$\% at L27.
  \item \textit{Per-feature (no anchor) collapses only at L27.} At L18 ($\|x\|/\sqrt{d} = 1.53$), the unconstrained per-feature variant survives with $0$\% dead and the best FVE ($0.726$). At L27 ($\|x\|/\sqrt{d} = 6.3$), it collapses to $83.4$\% dead. The cascade (\S\ref{app:perfeature-instability}) is gated by the norm-to-init ratio, not by absolute depth.
  \item \textit{Learned $a$ tracks the magnitude-as-noise gradient.} Global $a$ rises from $0.025$ (L9) to $0.257$ (L27); at L9 the optimizer wants pure cosine, at L27 it retains modest norm sensitivity for reconstruction. No value exceeds $0.3$; the inner-product regime ($a = 1$) is never approached.
\end{enumerate}

The depth story connects the mechanism (\S\ref{sec:exp-mechanism}) to the architecture (\S\ref{sec:ablation-matrix}): RMSNorm progressively erases magnitude from the residual stream at deeper layers, making inner-product scoring progressively worse as a feature detector. The cosine encoder's advantage grows with depth because the ``noise'' it removes (magnitude) becomes a larger fraction of the total signal at deeper layers. This explains why the headline result (L18, $500$M) is a moderate case; deeper layers show an even larger divergence between cosine and inner-product dictionaries, but at the cost of requiring more careful parameterization (base+delta rather than free per-feature).

\subsection{Cross-Budget Sparse Probing}
\label{app:cross-budget-probing}

Comparing our 50M-token cosine SAEs against the independently trained reference SAE \citep{karvonen2025saebench} at 500M tokens (same model, same BatchTopK recipe, different codebase):

\begin{table}[h]
  \centering
  \small
  \caption{Sparse-probing top-1 across token budgets. The cosine SAE at $10\times$ fewer tokens surpasses the standard reference at L9 and L27.\DIAG{exp51}{experiments/51\_saebench\_multilayer/README.md}}
  \label{tab:cross-budget-probing}
  \begin{tabular}{lccc}
    \toprule
    Layer & Cosine 50M & Standard 500M & $\Delta$ \\
    \midrule
    L9  & $\mathbf{0.786}$ & $0.613$ & $+17.3$\% \\
    L27 & $\mathbf{0.849}$ & $0.762$ & $+8.7$\% \\
    \bottomrule
  \end{tabular}
\end{table}

The reference SAE matches our own standard baseline within noise at L18 (both $\approx 0.679$ at 500M), confirming it is representative. The cross-budget gap is largest at L9, where the cosine encoder's direction-only scoring is most beneficial relative to the shallow-layer norm distribution.

\subsection{Sparsity-Budget Sweep}
\label{app:k-sweep}

We vary the BatchTopK sparsity budget $k$ (active features per token) at $50$M tokens, L18, $d_{\mathrm{sae}} = 65{,}536$, training Standard and Per-Feature Adaptive Cosine SAEs at $k \in \{10, 20, 40, 80, 160\}$ (the headline uses $k=80$).\DIAG{exp66}{experiments/66\_k\_sweep/README.md} Two findings (Table~\ref{tab:k-sweep}). First, the sparse-probing advantage is robust across the full budget range: the per-feature top-1 gain over Standard is $+8$ to $+11$\% at every $k$, with no collapse at low $k$ (it is in fact largest, $+11.4$\%, at $k=40$). Second, per-feature interpretability stays matched at every $k$: the describe-then-predict rate (\S\ref{app:interp-500m} protocol, $200$ features per arm) tracks between the two architectures within sampling noise, the arms trading the lead. Both architectures' interpretability rates decline as $k$ grows, since more simultaneously-active features are individually harder to characterize. The sparsity sweep thus reinforces the central decomposition: the cosine advantage is in feature \emph{discovery} (probing), not in per-feature interpretability, and this holds across sparsity budgets.

\begin{table}[h]
  \centering
  \small
  \caption{Sparsity-budget sweep (Qwen3-8B L18, $50$M tokens, $d_{\mathrm{sae}} = 65{,}536$). \emph{Top-1}: SAEBench single-feature probing. \emph{Interp}: describe-then-predict interpretable rate ($200$ features/arm). Std $=$ Standard, PF $=$ Per-Feature Adaptive Cosine.\DIAG{exp66}{experiments/66\_k\_sweep/README.md}}
  \label{tab:k-sweep}
  \begin{tabular}{lccccc}
    \toprule
    $k$ & 10 & 20 & 40 & 80 & 160 \\
    \midrule
    Top-1 Std    & $0.653$ & $0.695$ & $0.678$ & $0.731$ & $0.693$ \\
    Top-1 PF     & $\mathbf{0.733}$ & $\mathbf{0.782}$ & $\mathbf{0.792}$ & $\mathbf{0.816}$ & $\mathbf{0.788}$ \\
    \quad gap    & $+8.0$ & $+8.7$ & $+11.4$ & $+8.5$ & $+9.5$ \\
    \midrule
    Interp Std   & $0.530$ & $0.395$ & $0.350$ & $0.350$ & $0.180$ \\
    Interp PF    & $0.495$ & $0.440$ & $0.325$ & $0.300$ & $0.215$ \\
    \bottomrule
  \end{tabular}
\end{table}

\subsection{Cross-Model Summary}
\label{app:cross-model}

On Gemma-2-2B at 50M, $d_{\mathrm{sae}} = 9{,}216$: fixed-SAE top-1 sparse-probing difference $+3.4 \pm 0.3\%$.\DIAG{exp57c}{experiments/57c\_gemma\_replication/README.md} On Pythia-2.8B, Pythia-6.9B, Falcon-7B, the cosine top-1 advantage at deep layers does not hold; cos $>$ inner drops from $\sim 100\%$ at shallow to 40\% at deep (Pythia-2.8B L24 sharpest).\DIAG{exp25}{experiments/25\_multimodel\_matrix/README.md}

\textbf{Scaling with model size and expansion ratio.} A $3 \times 3$ matrix (Qwen3-1.7B/4B/8B $\times$ 4$\times$/8$\times$/16$\times$ expansion, 50M tokens, same recipe, three SAE-training seeds per cell) isolates the two potential drivers (Figure~\ref{fig:scaling-matrix}). Model size is the primary factor: the row mean rises from $+5.3$\% at $d_{\mathrm{model}} = 2048$ to $+11.7$\% at $d_{\mathrm{model}} = 2560$, then holds ($+9.2$\% at $d_{\mathrm{model}} = 4096$); the gain is large at every size but not strictly monotonic in $d_{\mathrm{model}}$. Expansion ratio has no systematic effect (column means $+9.4$/$+8.9$/$+7.9$\%, overlapping within seed noise). Per-cell seed SD is small ($\approx 1$--$3$\%). This revises the earlier interpretation that dictionary size drives the gap; the Gemma-2-2B result ($+3.4$\% at $d_{\mathrm{model}} = 2304$, 4$\times$ expansion) is explained by its small model dimension, not its small dictionary.\DIAG{exp67}{experiments/67\_scaling\_reseed/README.md}

\ifshownoaux
\subsection{FVE and Dead-Feature Trends Without Auxiliary Loss}
\label{app:fve-dead}

By-layer view of the no-auxiliary-loss runs cited in \S\ref{sec:exp-headline} (Qwen3-8B, 50M tokens, $d_{\mathrm{sae}} = 16{,}384$). FVE (Fig.~\ref{fig:fve-layer}, left) places Adaptive Cosine SAE and Per-Feature Adaptive Cosine SAE above Standard at L9 and L27, with the gap narrowing at L18. Dead-feature rate (right) separates the cosine variants at L18: Per-Feature stays below Standard, while the single-global-$a$ Adaptive variant matches Standard there and only recovers at L27. With the auxiliary loss restored, the dead-feature gap collapses to $1.9\%$ at L18 (500M tokens).

\begin{figure}[h!]
  \centering
  \includegraphics[width=0.46\linewidth,height=1.4in,keepaspectratio]{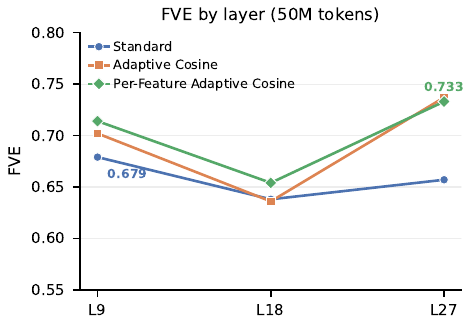}\hfill
  \includegraphics[width=0.46\linewidth,height=1.4in,keepaspectratio]{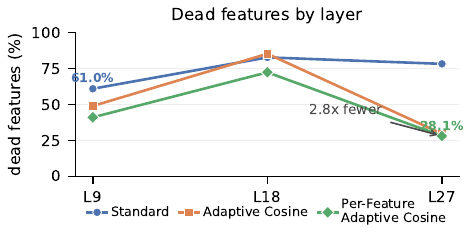}
  \caption{Without auxiliary loss, FVE and dead-feature rate by layer (Qwen3-8B, 50M tokens). At 500M tokens the L18 dead-feature gap (Per-Feature Adaptive Cosine SAE vs.\ Standard) is $1.9\%$.\DIAG{exp17}{experiments/17\_production\_scale/README.md}\DIAG{exp42c}{experiments/42c\_noc\_500m/README.md}}
  \label{fig:fve-layer}
  \label{fig:dead-by-layer}
\end{figure}
\fi

%% file: sections/06c-interpretability.tex
\section{Feature Interpretability}
\label{sec:interpretability}

\textbf{Setup.} We score features with an LLM judge under a describe-then-predict protocol: the judge writes a description from a feature's top activating contexts, then predicts which held-out tokens it fires on; ``interpretable'' means a feature is human-recognizable in the sense that its activations are predictable from a short natural-language description ($\geq 50$\% held-out prediction accuracy). At the 500M headline (Qwen3-8B L18, $1{,}000$ stratified features per architecture, Sonnet judge), per-feature interpretability is matched across architectures (Table~\ref{tab:interp-500m}, \S\ref{app:interp-500m}): $20.1\%$ (Standard), $21.3\%$ (global $a$), $19.2\%$ (per-feature). The same is true at $50$M / L27 under an alternative $1$--$5$ rubric ($40.0\%$ vs.\ $37.0\%$, $p = 0.88$).\DIAG{exp62}{experiments/62\_interp\_causal\_headline/README.md}\DIAG{exp33}{experiments/33\_feature\_interpretability/README.md} The cosine advantage is therefore in feature \emph{discovery} (\S\ref{sec:mechanism}), not in the legibility of individual features.

\begin{figure}[h]
  \centering
  \includegraphics[width=0.85\columnwidth,height=2.0in,keepaspectratio]{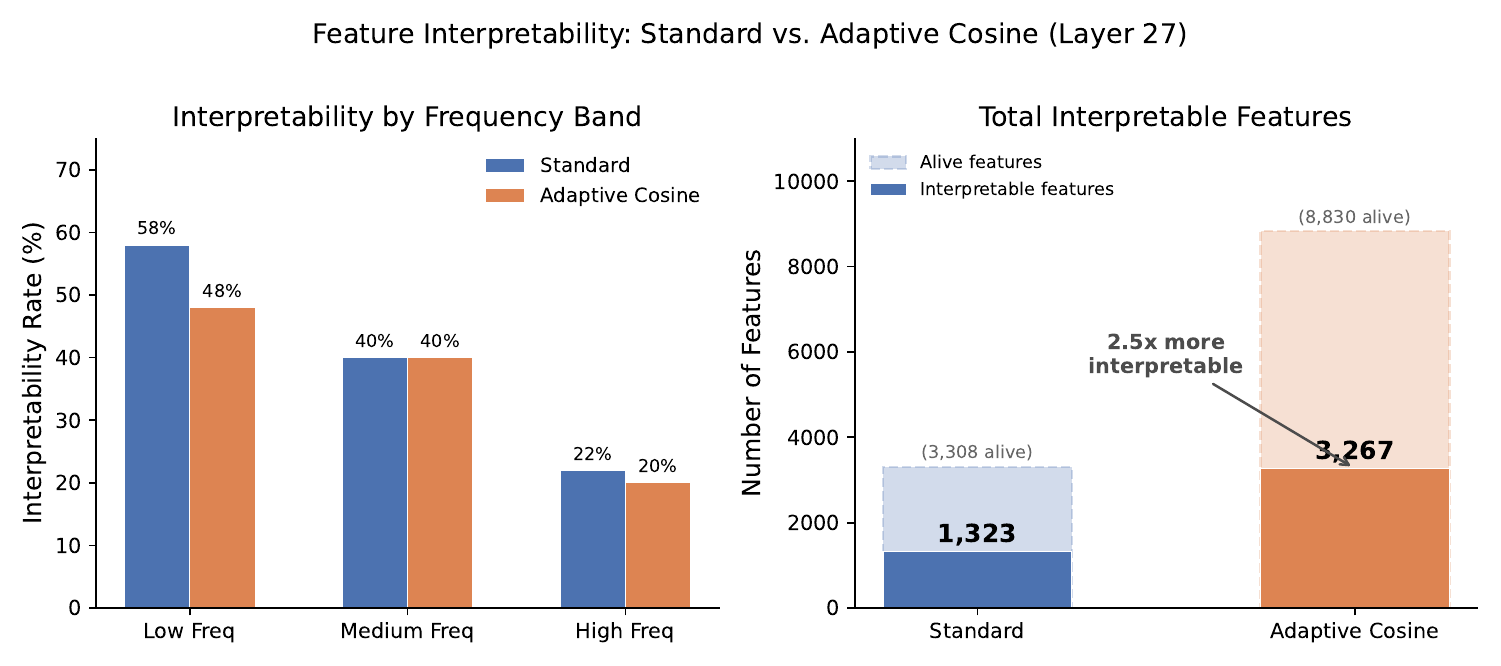}
  \caption{Per-feature interpretability rates are matched across architectures\ifshownoaux; when the alive-feature gap is open (no auxiliary loss), the cosine encoder nonetheless yields more interpretable features in total because more features are alive\fi.\DIAG{exp62}{experiments/62\_interp\_causal\_headline/README.md}}
  \label{fig:interpretability}
\end{figure}

\textbf{Decoder direction overlap.} $> 0.7$ cosine match: $31\%$ of $1000$ sampled features. Strict-identity overlap is $6$--$17\%$ (\S\ref{sec:exp-headline}).\DIAG{exp56b}{experiments/56b\_feature\_overlap/README.md}

\ifshownoaux
\subsection{Total Interpretable Features Without the Auxiliary Loss}
\label{app:interpretability}

Per-feature rates being matched, the count of interpretable features tracks how many features stay alive. With the auxiliary dead-feature loss (the deployed recipe), both architectures keep $\geq 50$k alive. Without it, the alive-feature gap reopens: at $100$M tokens, scoring $1{,}000$ stratified features on a $1$--$5$ rubric ($\geq 4$ interpretable), the norm-preserving cosine variant keeps far more features alive and therefore yields $\approx 4.5\times$ more interpretable features in total, at a matched per-feature rate (Table~\ref{tab:interp-summary}). This is the no-auxiliary-loss stress test of \S\ref{sec:aux-k}, not the headline recipe, and the arm shown is the norm-preserving (Magnitude-Bypass) variant.

\begin{table}[H]
  \centering
  \caption{LLM-judged interpretability without the auxiliary loss ($100$M tokens, $1{,}000$ stratified features, $1$--$5$ rubric, $\geq 4$ interpretable). Per-feature rates matched; the $\approx 4.5\times$ total-count gap comes from more features staying alive, not higher per-feature quality.\DIAG{exp40}{experiments/40\_saprmarks\_recipe/README.md}}
  \label{tab:interp-summary}
  \begin{tabular}{lccc}
    \toprule
    Variant ($100$M, no aux) & Per-feature $\geq 4$ & Alive & Total $\geq 4$ \\
    \midrule
    Standard                       & $82.1\%$ & $3{,}529$ & $\approx 2{,}900$ \\
    Magnitude-Bypass (norm-preserving) & $80.1\%$ & $16{,}332$ & $\approx \mathbf{13{,}100}$ \\
    \bottomrule
  \end{tabular}
\end{table}
\fi

\subsection{LLM Interpretability at 500M Tokens}
\label{app:interp-500m}

At $500$M tokens with the production recipe (matched alive counts via aux-k), LLM-judged describe-then-predict rates are matched across architectures: $1{,}000$ stratified features per arm, scored by a Sonnet judge, fall within a $2.1$-point band (Table~\ref{tab:interp-500m}). Per-feature cosine is statistically indistinguishable from standard ($19.2\%$ vs.\ $20.1\%$), with global cosine marginally highest ($21.3\%$). Alive counts are matched ($17.3$--$18.7$k), so this is a like-for-like per-feature comparison.

\begin{table}[h]
  \centering
  \small
  \caption{LLM-judged interpretability at $500$M tokens, $1{,}000$ stratified features per architecture, Sonnet judge. Interpretable iff $\geq 50$\% prediction accuracy. Per-feature rates are matched; the cosine advantage is in feature discovery (\S\ref{sec:mechanism}), not per-feature interpretability.\DIAG{exp62}{experiments/62\_interp\_causal\_headline/README.md}}
  \label{tab:interp-500m}
  \begin{tabular}{lcccc}
    \toprule
    Variant & Rate & Low-freq & Med-freq & High-freq \\
    \midrule
    Standard        & $20.1\%$ & $22.3\%$ & $21.6\%$ & $14.8\%$ \\
    Global $a$      & $\mathbf{21.3\%}$ & $24.3\%$ & $21.4\%$ & $18.0\%$ \\
    Per-feature     & $19.2\%$ & $23.5\%$ & $19.5\%$ & $14.3\%$ \\
    \bottomrule
  \end{tabular}
\end{table}

This is the expected reading given the rest of the analysis: per-feature interpretability, like per-feature steering power (\S\ref{sec:mechanism}), is matched across architectures, and the cosine advantage lives in which features the dictionary discovers rather than in the legibility of any individual feature. A smaller $200$-feature pilot (\citealt{karvonen2025saebench}-style stratification) suggested a frequency crossover, with per-feature cosine ahead on low-frequency features; that pattern does not survive at $1{,}000$ features (low-frequency rates fall within $2$ points across arms), so we attribute it to small-sample noise and do not claim a frequency-dependent interpretability effect.

\subsection{Concrete Feature Examples}
\label{app:qualitative-features}

Table~\ref{tab:feature-examples} makes ``interpretable'' concrete: for each architecture we show one feature the judge could describe and then use to predict held-out firing (accuracy $1.0$), and one it could not ($\leq 0.1$). The interpretable features are crisp lexical or syntactic detectors under every score (e.g.\ the word ``you,'' the adverbial ``-ly'' suffix, ``the'' following ``and''); the uninterpretable ones fire on heterogeneous token sets with no description that predicts them (e.g.\ subword fragments completing a split name). Both dictionaries contain features of each kind, which is why per-feature rates are matched (Table~\ref{tab:interp-500m}); consistent with the salient-concept control (\S\ref{app:vignette}), the cosine advantage is distributional, not a matter of individual cosine features being more legible.

\begin{table}[h]
  \centering
  \small
  \caption{Representative interpretable (prediction accuracy $1.0$) and uninterpretable ($\leq 0.1$) features per architecture at the $500$M headline, with the LLM-written description and top activating tokens. Both architectures produce clean and fragmentary features alike.\DIAG{exp62}{experiments/62\_interp\_causal\_headline/README.md}}
  \label{tab:feature-examples}
  \begin{tabular}{p{0.16\columnwidth}cp{0.42\columnwidth}c}
    \toprule
    Architecture & Feat. & Auto-interp description (abbrev.) & Acc. \\
    \midrule
    Standard       & $37058$ & word ``you'' (subject pronoun) & $1.0$ \\
    Standard       & $16263$ & head noun following a modifier (no predictive pattern) & $0.1$ \\
    \midrule
    Global $a$     & $52155$ & adverbial ``-ly'' suffix & $1.0$ \\
    Global $a$     & $37895$ & mixed post-verb temporal tokens (no predictive pattern) & $0.0$ \\
    \midrule
    Per-feature    & $19859$ & ``the'' immediately after ``and'' & $1.0$ \\
    Per-feature    & $33290$ & subword fragment completing a split first name & $0.1$ \\
    \bottomrule
  \end{tabular}
\end{table}

%% file: sections/07-gradient-equalization.tex
\section{Mechanism: Gradient Equalization and Q4 Reconstruction}
\label{sec:mechanism}

This section uses the three cosine variants from \S\ref{sec:method}: \emph{global $a$} (single learned exponent), \emph{per-feature} ($a_i = a_{\mathrm{base}} + \delta_i$), and \emph{magnitude-bypass} (pinned $a{=}0$ with norm restoration; Appendix~\ref{sec:cosine-sae}).

\textbf{Encoder-gradient asymmetry.} For an active feature under a fixed BatchTopK/ReLU mask, the inner-product score gradient is $\partial s_i / \partial w_i = x_c$, so the encoder update inherits a factor of $\|x_c\|$. Cosine scoring replaces this with a normalized direction times a scalar $\|x_c\|^a$; the trained $a \ll 1$ (Table~\ref{tab:scale-a}).

\begin{figure}[h]
  \centering
  \includegraphics[width=0.85\columnwidth,height=1.8in,keepaspectratio]{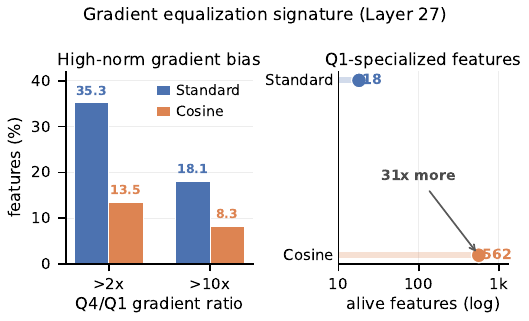}
  \caption{Encoder-gradient ratio Q4 (high-norm) / Q1 (low-norm). Standard: $35.3\%$ of features have Q4/Q1 $> 2$. Per-feature cosine: $13.5\%$.\DIAG{exp28}{experiments/28\_gradient\_analysis/README.md}}
  \label{fig:gradient-dominance}
\end{figure}

\textbf{Equalization signature.} Cosine scoring compresses the gradient ratio toward unity. The median per-feature Q4/Q1 ratio drops from $1.55\times$ (Standard) to $1.03\times$ (per-feature cosine). Only $13.5\%$ of cosine features exceed Q4/Q1 $> 2$, versus $35.3\%$ under standard scoring. Q1-specialized features increase from $18$ to $562$ (Fig.~\ref{fig:gradient-dominance}).\DIAG{exp28}{experiments/28\_gradient\_analysis/README.md}

\textbf{Recipe and dictionary-size context.} On Gemma-2-2B ($d_{\mathrm{sae}} = 9{,}216$, 4$\times$ expansion, 50M tokens), the community SAEBench recipe's auxiliary loss eliminates dead features for both architectures, narrowing the cosine advantage to $+3.4$\%. The scaling matrix (\S\ref{app:cross-model}) shows that model dimension, not expansion ratio, drives the gap; the headline $+14.6$\% emerges at Qwen3-8B's $d_{\mathrm{model}} = 4096$.\DIAG{exp57c}{experiments/57c\_gemma\_replication/README.md}\DIAG{exp57}{experiments/57\_scaling\_matrix/README.md}

\textbf{Discovery vs.\ separability.} How much of the probing gap comes from features the cosine encoder discovers versus cleaner encoding of shared features? We restrict both 500M dictionaries to $8{,}661$ mutually matched features (activation correlation $\geq 0.7$ and decoder cosine $> 0.7$). On matched features alone, top-1 difference shrinks from $+14.87\%$ to $+1.95\%$; top-5 from $+10.56\%$ to $+2.70\%$. The standard SAE's unmatched features contribute nothing beyond its matched set (full $-$ matched at top-5: $0.7831 - 0.7837$). The cosine SAE's unmatched features add $+7.8\%$ ($0.8888 - 0.8108$).\DIAG{exp58b}{experiments/58b\_matched\_feature\_probing/README.md}

The advantage is therefore distributional, not a failure of standard scoring to represent salient concepts. For hand-picked content concepts (a programming language, a natural language), \emph{both} dictionaries contain a cleanly selective feature: the top concept-aligned feature fires strongly on-concept and is silent off-concept under either score (Appendix~\ref{app:vignette}). The gap lives in the aggregate over the eight probing categories and the long tail of features, consistent with the matched-feature decomposition above, rather than in any single obvious concept.\DIAG{exp65}{experiments/65\_causal\_vignette/README.md}

\textbf{Unmatched features by norm quartile.} Where do each model's unique features fire? Among strict-unmatched features at L18, $86.3\%$ of Standard activations fall in Q4 vs.\ $56.9\%$ for per-feature cosine. The mean norm of Standard-unique activations is $9{,}689$ ($48\times$ the sampled token mean). On tokens where per-feature cosine-unique features fire, Standard fires $328$ features/token vs.\ $122$ for per-feature cosine.\DIAG{exp58c}{experiments/58c\_norm\_detector\_characterization/README.md}

\textbf{Decoder geometry control.} For ten cached TPP probe directions, $w_{\mathrm{probe}} W_{\mathrm{dec}}^\top$ has matched concentration: entropy ratio $1.00001$, effective-dimension ratio $1.00007$.\DIAG{exp58a}{experiments/58a\_probe\_weight\_analysis/README.md}

\textbf{Feature steering behavioral equivalence.} To verify that the sparse-probing gap reflects feature \emph{discovery}, not per-feature \emph{power}, we directly steer individual features at amplification factors $0.5\times$, $2\times$, and $5\times$ and measure downstream KL-divergence. Across $11$ top-activating standard features and $9$ cosine features ($50$ prompts each), mean KL ratios are $1.04\times$ ($0.5\times$), $1.00\times$ ($2\times$), and $1.00\times$ ($5\times$): cosine and standard features produce indistinguishable behavioral effects at matched amplification. This confirms that the $+14.9$\% probing gap is not explained by standard features being individually ``stronger''; rather, cosine discovers more concept-aligned directions.\DIAG{exp55d}{experiments/55d\_feature\_steering/README.md}

\textbf{Causal ablation cleanliness.} Sparse probing is a proxy; we therefore test the features causally. Ablating the top-$N$ probe-selected features and decomposing the logit change into \emph{intended} (target-class) and \emph{unintended} (collateral) effects, the cosine dictionary ablates far more cleanly. The precision ratio (intended/unintended effect) rises with $N$ for cosine, from $8.4\times$ at $N{=}2$ to $21.8\times$ at $N{=}500$, while standard \emph{degrades} from $5.6\times$ to $1.4\times$ (Table~\ref{tab:tpp-precision}). The gain is driven by the denominator, not the numerator: at $N{=}500$ the intended effect is comparable across architectures ($0.44$ Standard vs.\ $0.37$ cosine), but collateral damage stays nearly flat for cosine ($0.0007 \to 0.017$) while growing two orders of magnitude for standard ($0.0028 \to 0.30$). The cleanliness is therefore a property of \emph{which} features the dictionary holds, not of any per-feature separability advantage: it is the norm-conditioned features (\S\ref{sec:mechanism}), absent from the cosine dictionary, that bleed diffuse side effects when ablated as a set. This is consistent with the matched single-feature steering above and with the small ($+2.0$\%) separability component of the probing gap; the cosine dictionary spends capacity on content rather than norm, now established through intervention rather than a linear proxy.\DIAG{exp57b}{experiments/57b\_tpp\_collateral/README.md}

\begin{table}[h]
  \centering
  \small
  \caption{\textbf{Ablation precision (intended/unintended logit effect) vs.\ number of ablated features.} Numeric form of Fig.~\ref{fig:tpp-precision}. Higher is cleaner. Cosine sharpens with $N$; standard collapses toward $1\times$ as collateral (not weaker intended effect) overwhelms it. Intended effects are comparable across architectures; the gap is in collateral. Qwen3-8B L18, 500M.\DIAG{exp57b}{experiments/57b\_tpp\_collateral/README.md}}
  \label{tab:tpp-precision}
  \begin{tabular}{lccccc}
    \toprule
    $N$ ablated & $2$ & $10$ & $50$ & $100$ & $500$ \\
    \midrule
    Standard       & $5.6\times$ & $5.9\times$ & $4.7\times$ & $2.5\times$ & $1.4\times$ \\
    Cosine         & $\mathbf{8.4\times}$ & $\mathbf{9.5\times}$ & $\mathbf{14.1\times}$ & $\mathbf{13.9\times}$ & $\mathbf{21.8\times}$ \\
    \bottomrule
  \end{tabular}
\end{table}

\begin{figure}[h]
  \centering
  \includegraphics[width=0.78\columnwidth,height=2.2in,keepaspectratio]{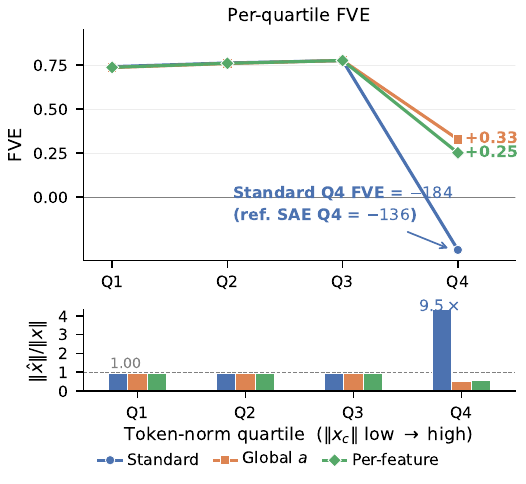}
  \caption{Per-quartile FVE (top) and reconstruction-norm ratio $\|\hat x\| / \|x\|$ (bottom).\DIAG{exp55c}{experiments/55c\_norm\_stratified\_reference/README.md}}
  \label{fig:q4-app}
\end{figure}

\begin{table}[h]
  \centering
  \small
  \caption{FVE by activation-norm quartile (Q1 lowest, Q4 highest).\DIAG{exp55c}{experiments/55c\_norm\_stratified\_reference/README.md}}
  \label{tab:q4-stratified}
  \begin{tabular}{lcccc}
    \toprule
        & Std.\ & Global $a$ & Per-feat.\ & Mag.-byp.\ \\
    \midrule
    Q1  & $0.738$ & $0.732$ & $0.733$ & $0.738$ \\
    Q2  & $0.763$ & $0.760$ & $0.762$ & $0.762$ \\
    Q3  & $0.778$ & $0.777$ & $0.778$ & $0.775$ \\
    Q4  & $-183.5$ & $0.329$ & $0.252$ & $-0.200$ \\
    \bottomrule
  \end{tabular}
\end{table}

\textbf{Q4 reconstruction.} Q1--Q3 FVE is matched; Q4 splits: Standard $-183.5$, global cosine $0.33$, per-feature cosine $0.25$, magnitude-bypass $-0.20$. Reconstruction-norm ratio (output-norm / input-norm) on Q4: Standard $9.5\times$; cosine variants $\approx 1\times$.\DIAG{exp55c}{experiments/55c\_norm\_stratified\_reference/README.md}

\textbf{Controls.} Can cosine scoring be applied post-hoc to a trained Standard SAE? No: FVE drops $-18$ to $-33\%$, $L_0$ jumps from 80 to 500, and $< 11\%$ of features overlap with from-scratch cosine (\S\ref{app:posthoc}).\DIAG{exp29}{experiments/29\_posthoc\_normalization/README.md}\DIAG{exp17}{experiments/17\_production\_scale/README.md} Input normalization without unit-norm encoder rows kills $33\%$ of alive features.\DIAG{exp45}{experiments/45\_encoder\_norm\_ablation/README.md} Global cosine has higher FVE than Standard at every learning rate tested.\DIAG{exp30}{experiments/30\_lr\_sweep/README.md}

\begin{figure}[h]
  \centering
  \includegraphics[width=\columnwidth]{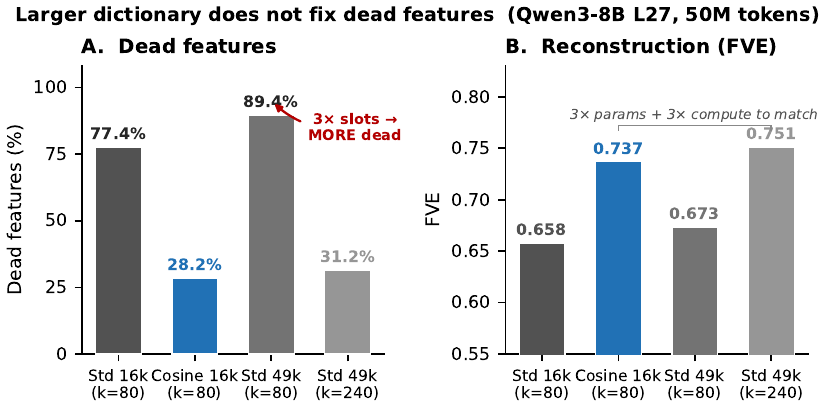}
  \caption{\textbf{Larger dictionary does not fix dead features.} Qwen3-8B L27, 50M tokens. \textbf{(A)} Tripling dictionary slots at the same L0 increases the dead rate from $77.4$\% to $89.4$\%; the cosine SAE at $1/3$ the dictionary achieves $28.2$\%. \textbf{(B)} Matching cosine's FVE requires $3\times$ parameters and $3\times$ L0. The dead-feature problem is a scoring-function pathology, not a capacity limitation.\DIAG{exp26}{experiments/26\_dictionary\_size/README.md}}
  \label{fig:dict-size-control}
\end{figure}

\textbf{Dictionary size control.} Giving the standard encoder $3\times$ more dictionary slots at matched L0 makes dead rates \emph{worse}, not better: $89.4\%$ dead (49k, $k{=}80$) vs.\ $77.4\%$ (16k, $k{=}80$), while a 16k global cosine achieves $28.2\%$ dead (Figure~\ref{fig:dict-size-control}). The gradient concentration is a property of the scoring function, not dictionary capacity. With $3\times$ dictionary \emph{and} $3\times$ L0 ($k{=}240$, $3\times$ parameters, $3\times$ compute), the standard SAE can slightly exceed the cosine SAE's FVE ($0.751$ vs.\ $0.737$), confirming that cosine is $\sim 3\times$ more parameter-efficient.\DIAG{exp26}{experiments/26\_dictionary\_size/README.md}

\textbf{Cos $>$ inner across architectures.} Which variant actually shows a cos$>$inner metric difference? At L18 (community recipe), only magnitude-bypass reaches $69\%$. global cosine ($a = 0.258$), per-feature cosine (mean $a_i = 0.076$), and Standard all cluster at $62$--$63\%$. Only the architecture that removes magnitude entirely from $z$ separates on this metric.\DIAG{exp44}{experiments/44\_norm\_stratified\_fve/README.md}

\textbf{Architectures and the cos-vs-inner difference.} All cosine variants improve sparse probing: magnitude-bypass $+11.6\%$ top-1, global cosine $+13.3\%$, per-feature cosine $+14.9\%$. Only magnitude-bypass shows the cos-vs-inner metric difference. The RMS-geometric-alignment mechanism is depth-dependent; the gradient-equalization signature appears for all cosine variants. At L9, cos $>$ inner falls below 50\%, yet both cosine variants still improve sparse probing.\DIAG{exp40/42c/44}{experiments/40\_saprmarks\_recipe/README.md}

\subsection{Norm-Stratified FVE}
\label{app:q4}

Per-quartile values are reported once in Table~\ref{tab:q4-stratified} above. Reconstruction-norm ratio on Q4 tokens: Standard $9.5\times$; cosine variants $\approx 1.0\times$. The reference SAE of \citet{karvonen2025saebench} reproduces the pattern with Q4 FVE $= -136$.\DIAG{exp55c}{experiments/55c\_norm\_stratified\_reference/README.md}

\subsection{Post-Hoc Normalization and Score Swapping}
\label{app:posthoc}

Cosine scoring applied at inference time to a trained Standard SAE: FVE $-18$ to $-33\%$, $L_0$ from $80$ to $500$, Jaccard with from-scratch global cosine $10.6\%$.\DIAG{exp29}{experiments/29\_posthoc\_normalization/README.md}

\paragraph{Direct score swap (both directions).} We train a Standard and an adaptive-cosine BatchTopK SAE at the headline setting (Qwen3-8B L18, $50$M tokens, $d_{\mathrm{sae}} = 65{,}536$, $k = 80$) and swap the score geometry at inference: the same weights are read with the other encoder (the adaptive-cosine score recovers inner product at $a = 1$). This tests the mechanism more directly than gradient reweighting (\S\ref{app:q4}): does the score, holding weights fixed, move the result? Table~\ref{tab:score-swap} separates two effects.\DIAG{exp64}{experiments/64\_score\_swap/README.md}

\begin{table}[h]
  \centering
  \small
  \caption{\textbf{Score swap at inference: FVE is training-bound, probing is read-time.} Same checkpoint, scored two ways. Swapping the score destroys FVE in \emph{both} directions (reconstruction is fixed at training time), but sparse-probing top-1 tracks the \emph{score used at read time}, not the score the dictionary was trained with.\DIAG{exp64}{experiments/64\_score\_swap/README.md}}
  \label{tab:score-swap}
  \begin{tabular}{llcc}
    \toprule
    Checkpoint & Scored as & FVE & Probe top-1 \\
    \midrule
    Standard & inner (native) & $0.709$ & $0.694$ \\
    Standard & cosine (swap)  & $0.599$ & $\mathbf{0.763}$ \\
    \midrule
    Cosine   & cosine (native) & $0.707$ & $0.766$ \\
    Cosine   & inner (swap)    & $0.612$ & $\mathbf{0.698}$ \\
    \bottomrule
  \end{tabular}
\end{table}

FVE falls by $\approx 0.11$ under either swap: a dictionary trained for one geometry reconstructs poorly when read with the other, confirming reconstruction quality is set during training. Sparse probing behaves oppositely: Standard weights read with the cosine score reach $0.763$ (matching the from-scratch cosine SAE's $0.766$), and cosine weights read with inner product fall to $0.698$ (matching from-scratch Standard's $0.694$). Probing quality therefore follows whichever geometry reads the dictionary, consistent with the decomposition of \S\ref{app:selectors}: the input-norm channel that governs probing is applied at read time, whereas the encoder weights (and their FVE) are fixed.

This is not a post-hoc shortcut. Reading a Standard dictionary with the cosine score recovers the probing number only by moving off its trained operating point: FVE drops to $0.599$ and $L_0$ drifts to $55$ (Table~\ref{tab:score-swap}), so the result is a worse autoencoder that happens to select more interpretable features, not a usable SAE. A deployable dictionary needs reconstruction \emph{and} interpretable selection at a matched sparsity, and only end-to-end cosine training delivers both; the swap isolates \emph{why} (the score sets selection) without substituting for training (the decoder must be fit under that score). This is the inference-time analogue of the destructive post-hoc result above and of exp29.

\paragraph{Continued-training swap.} Swapping the score and continuing training for $10$M tokens recovers FVE (to $\approx 0.71$) but does not reorganize the dictionary: the alive-feature set is nearly unchanged (Jaccard $\geq 0.998$ with the pre-swap dictionary in both directions) and decoder directions barely move (mean cosine $0.97$--$0.98$). At this budget the swap re-fits magnitudes and thresholds rather than re-sorting features; whether a longer continued-training budget would induce reorganization is left open.\DIAG{exp64}{experiments/64\_score\_swap/README.md}

\subsection{Per-Feature Gradient Ratio}
\label{app:grad-ratio}

\begin{figure}[h]
  \centering
  \includegraphics[width=\columnwidth]{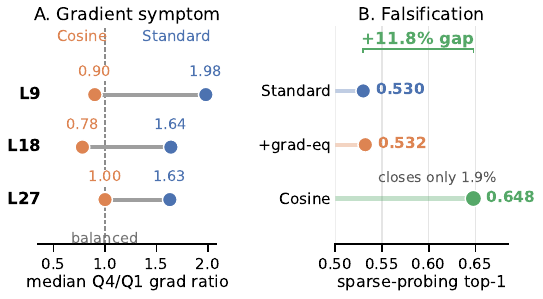}
  \caption{\textbf{Gradient asymmetry exists, but does not explain the probing gap.} \textbf{(A)} Median per-feature Q4/Q1 gradient ratio across Qwen layers (10M tokens). \textbf{(B)} Equalizing per-quartile gradients in the standard encoder by reweighting the reconstruction loss ($1/\|x_c\|$; 50M tokens) negligibly shifts top-1. Per-feature ratio definition and full reweighting sweep in text below.}
  \label{fig:grad-equalization}
\end{figure}

For each alive feature $i$: average $\|\nabla_{w_i}\mathcal{L}\|$ over tokens in each $\|x_c\|$ quartile, take Q4-to-Q1 ratio, report median across alive features (Fig.~\ref{fig:grad-equalization}). At Qwen L9/L18/L27 with 10M tokens: median Q4/Q1 is $1.6$--$2.0\times$ (Standard) vs.\ $0.8$--$1.0\times$ (global cosine); Q1-specialized features grow $\sim 4\times$ under cosine.\DIAG{exp54}{experiments/54\_gradient\_saprmarks/README.md}

\paragraph{Reweighting test.} If gradient asymmetry were the primary cause, equalizing it should close the gap. Reweighting the Standard reconstruction loss equalizes per-quartile gradient contributions by construction, without modifying the score. We test mild reweighting ($1/\|x_c\|$) and strong reweighting ($1/\|x_c\|^2$), spanning partial to near-complete cancellation of the encoder-gradient norm factor. At 50M tokens with FVE $\geq 0.97$, per-feature cosine reaches top-1 $0.648$. Standard moves from $0.530$ (no reweighting) to $0.532$ (mild) and $0.536$ (strong). Across $k \in \{1, 2, 5\}$, reweighting closes only $1.9$--$6.8\%$ of the cosine difference. The forward pass still inflates every pre-activation on Q4 tokens, BatchTopK still selects the same features, and the same norm-conditioned firing pattern reappears.\DIAG{exp59}{experiments/59\_gradient\_equalization/README.md}

%% file: sections/08-limitations.tex
\section{Limitations and Scope}
\label{sec:limitations}

\textbf{Scope.} All results are on the residual stream; no MLP-internal, attention-internal, head-output, or pre-norm activations. Causal diagnostics use greedy decoding and deterministic activation collection. Multi-layer / 50M+ runs cover Qwen3-8B fully; Gemma-2-2B partially; Mistral-7B after the init fix; Pythia and Falcon at 5M. No model is $> 8$B; no QK-norm or per-head-norm models. Training data: FineWeb (English-dominant). SAEBench is English-only (europarl is English-vs-other classification, not multilingual concept probing). No claim about multilingual residual streams, vision, audio, multimodal, or instruction-tuned / RLHF models.

\textbf{Where the metric difference weakens.} At Qwen L9, cos $>$ inner is below $50\%$. Deep LayerNorm: cos $>$ inner drops from $100\%$ to $40\%$ on Pythia-2.8B between L8 and L24 (inner-to-KL correlation $0.441$ vs.\ cos-to-KL $0.271$); Gemma-2-2B L20 (RMSNorm) is at $\approx 53\%$. Within-norm-quartile rates on Qwen are $40$--$70\%$ (Simpson's paradox; the 70--90\% rate is partly between-quartile). The architectural FVE / alive-feature differences persist across these settings.\DIAG{exp25}{experiments/25\_multimodel\_matrix/README.md}

\textbf{Architecture-specific failure modes.} \ifshownoaux Without auxiliary loss: Adaptive Cosine SAE retains $3.3\times$ alive-feature ratio over Standard. With auxiliary loss: both architectures reach $\sim 0\%$ dead. \fi Per-Feature Adaptive Cosine SAE: $83\%$ dead at 50M / L27 ($65$k independent scale parameters do not converge under low budget at high norms). Magnitude-Bypass SAE: $4.3\%$ persistent dead at 500M (auxiliary-loss gradients $\sim 6$ orders weaker through bounded $[0, 1]$ activations; \S\ref{app:norm-restore}). At 5M the auxiliary loss is inert for Magnitude-Bypass SAE (\S\ref{sec:exp-mechanism}). Adaptive Cosine SAE is the architecture without these failure modes in our coverage.

\textbf{Initialization.} Mistral has $\|x\| \approx 6$ vs.\ Qwen $\approx 400$. $\sqrt{d}$ init on Mistral: $> 95\%$ dead. Norm-adaptive init helps at 5M but reduces FVE at 50M+ on Qwen. Default: $\sqrt{d}$ when $\|x\|$ is within $2\times$ of $\sqrt{d}$, else $\log(\operatorname{mean}\|x\|)$ (\S\ref{app:init}).\DIAG{exp42}{experiments/42\_mistral\_init/README.md}

\textbf{Evaluation scope.} The 500M Qwen L18 headline reproduces across three SAE-training seeds (Appendix~\ref{app:stats}); per-dataset breakdowns and the Magnitude-Bypass arm remain single-seed. Cross-layer and cross-model SAEBench replications are pending (\S\ref{app:open-gaps}). No feature steering, circuit-level interventions, or downstream task accuracy. Sentiment is the only top-1 reversal in Table~\ref{tab:per-dataset-probing} (Standard $0.915$ vs.\ Per-Feature Adaptive Cosine SAE $0.880$).

\textbf{Decoder under norm shift.} Under norm-noise training (Uniform$(0.5, 2.0)$), cosine FVE drops $4.5$--$5.8\%$ vs.\ $1.8$--$3.4\%$ for Standard.\DIAG{exp31}{experiments/31\_norm\_injection/README.md} The encoder is norm-invariant; the decoder reconstructs $x$ and depends on norm. Where train and eval norms diverge, the headline difference may erode. \emph{adaptive\_l2} achieves FVE $0.738$; postnorm-loss achieves SAE $\to$ KL $0.380$; postnorm forces $a \to -0.4$.\DIAG{exp22}{experiments/22\_postnorm\_scale/README.md}\DIAG{exp14}{experiments/14\_postnorm\_loss/README.md}

\textbf{Architectures discover different features.} Strict-identity overlap of alive features between Standard and Per-Feature Adaptive Cosine SAE at L18: $< 1\%$ at $n = 3$--$6$ paired, consistent with the non-canonical nature of SAE dictionaries \cite{leask2025canonical}. Headline aggregate comparisons are across disjoint feature sets. Inner-product ablation $(x \cdot f) f$ scales with $\|x\|$, biasing the comparison toward Standard; under norm-invariant ablation, the Standard SAE $\to$ KL difference shrinks by $70\%$ at L27.\DIAG{exp56b}{experiments/56b\_feature\_overlap/README.md}

\subsection{Open Gaps}
\label{app:open-gaps}

\textbf{Coverage gaps.} Whether Standard's dead features overlap the cosine-only sparse-probing features is not known. LLM-judged interpretability ran at L27 only; L9 / L18 untested. No 8B+ LayerNorm model at 50M+ tokens. Cross-architecture coverage stops at 5M for some models.

\textbf{Production-scale variance.} Gradient-equalization measurements are at 5M; 500M re-measurement is pending.

\textbf{Architecture-design open questions.} Per-Feature Adaptive Cosine SAE vs.\ Magnitude-Bypass SAE as the preferred magnitude-stripping architecture is unsettled. The base$+\delta$ parameterization as a fix for Per-Feature Adaptive Cosine SAE's 50M dead-feature regime has not been validated at scale. No mechanistic account exists for why Per-Feature Adaptive Cosine SAE leads at top-1 while Adaptive Cosine SAE leads at top-5.

\textbf{Beyond sparse probing.} OOD shift and stochastic decoding (high temperature, nucleus) are not tested.

%% file: sections/09-discussion.tex
\section{Extended Discussion}
\label{sec:discussion}

\textbf{Triangulation across architectures.} Magnitude-Bypass SAE ($+11.6\%$ top-1, single seed), Adaptive Cosine SAE ($+14.1\%$), and Per-Feature Adaptive Cosine SAE ($+14.6\%$) all gain at matched FVE (\S\ref{sec:exp-headline}).\DIAG{exp40/42c/44}{experiments/40\_saprmarks\_recipe/README.md} Fitted $a = 0.258$ at the headline setting; $a_i \in [-0.5, 0.5]$ across $65{,}536$ features (Table~\ref{tab:scale-a}).\DIAG{exp42c}{experiments/42c\_noc\_500m/README.md}

\textbf{Mechanism dissociation.} The cos-vs-inner score advantage requires fully removing magnitude (Magnitude-Bypass SAE only); the gradient-equalization signature appears for all cosine variants (\S\ref{sec:exp-mechanism}). Cos~$>$~inner is below $50\%$ at L9, yet both cosine probes still improve sparse probing at L9.

\textbf{Open questions.} Can magnitude be retained where informative (e.g.\ sentiment) without dominating feature detection? Do the dead features of the standard SAE correspond to the cosine features that drive sparse probing? Decoder-cosine overlap between cosine and standard dictionaries is $6$--$17\%$ at the $> 0.95$ threshold (\S\ref{sec:exp-headline}, consistent with \citet{leask2025canonical}); a stitching or meta-SAE analysis across cosine seeds is needed to distinguish stable-target densification from dictionary resampling.\DIAG{exp56b}{experiments/56b\_feature\_overlap/README.md}

%% file: sections/appendix.tex
\section{Training Recipe}
\label{app:recipe}

Training uses the Marks-lab \texttt{dictionary\_learning} library (\url{https://github.com/saprmarks/dictionary_learning}) accessed via \texttt{sae-bench}~\cite{karvonen2025saebench}. The recipe matches OpenAI's TopK SAE training~\cite{gao2024scaling} except for the BatchTopK selection rule~\cite{bussmann2024batchtopk} and the encoder-score swap (\S\ref{sec:cosine-sae}).

\begin{table}[H]
  \centering
  \caption{Headline training recipe (Qwen3-8B L18, 500M tokens). All four architectures (Standard, Adaptive Cosine SAE, Per-Feature Adaptive Cosine SAE, Magnitude-Bypass SAE) share these settings. The Per-Feature Adaptive Cosine headline runs use the training/eval code in exp42c.\DIAG{exp40}{experiments/40\_saprmarks\_recipe/README.md}\DIAG{exp42c}{experiments/42c\_noc\_500m/README.md}}
  \label{tab:recipe-headline}
  \begin{tabular}{ll}
    \toprule
    Setting & Value \\
    \midrule
    Model / site                  & \texttt{Qwen/Qwen3-8B-Base}, residual, layer 18 \\
    Activation / dictionary dim   & $d_{\mathrm{model}} = 4096$, $d_{\mathrm{sae}} = 65{,}536$ \\
    Tokens / steps                & $500$M tokens, $244{,}140$ steps \\
    Batch / sparsity              & batch $2048$, BatchTopK $k = 80$ \\
    Optimizer / LR                & Adam, LR $5 \cdot 10^{-5}$, seed $42$ \\
    LR schedule                   & 1000-step warmup; constant until 80\%, then linear decay \\
    Auxiliary loss                & weight $1/32$, $k_{\mathrm{aux}} = 2048$, dead threshold $10$M tokens \\
    Decoder                       & unit-norm decoder rows after each step; geometric-median $b_{\mathrm{dec}}$ \\
    Stability                     & gradient clipping max norm $1.0$; bf16 activations \\
    Eval cache                    & $2$M held-out tokens \\
    \bottomrule
  \end{tabular}
\end{table}

\textbf{Data.} Training tokens come from the FineWeb \texttt{sample-10BT} subset (\texttt{HuggingFaceFW/fineweb}), tokenized with each model's bundled tokenizer. SAEBench probing datasets are at the versions shipped with \texttt{sae-bench}.

\textbf{Reproducibility.} Code and full experiment scripts are available at \url{https://github.com/SilenNaihin/cosine-scored-saes}. The 500M-token headline SAE checkpoints (Standard, Global $a$, Per-Feature Adaptive Cosine) are released at \url{https://huggingface.co/Silen/cosine-scored-saes-qwen3-8b}.

\section{Statistical Summary}
\label{app:stats}

\begin{table}[H]
  \centering
  \small
  \caption{Effect sizes and significance.}
  \label{tab:effect-sizes}
  \begin{tabular}{p{0.62\linewidth}p{0.30\linewidth}}
    \toprule
    Claim & Strength \\
    \midrule
    \ifshownoaux FVE difference $+8.0\%$ at L27, 50M tokens, no auxiliary loss\DIAG{exp34}{experiments/34\_multi\_seed/README.md} & $41.7\sigma$, $n = 3$ seeds \\ \fi
    Dead-feature difference $48.7\%$\DIAG{exp17}{experiments/17\_production\_scale/README.md} & $37.5\sigma$ \\
    Cos $>$ inner above $50\%$ at L27\DIAG{exp25}{experiments/25\_multimodel\_matrix/README.md} & $p < 2.4 \times 10^{-6}$ \\
    Dictionary efficiency: 16k Adaptive Cosine SAE alive $>$ 49k Standard alive\DIAG{exp17}{experiments/17\_production\_scale/README.md} & Cohen's $h = 1.358$ \\
    Gradient Q4 dominance: $35.3\%$ vs.\ $13.5\%$\DIAG{exp28}{experiments/28\_gradient\_analysis/README.md} & Cohen's $h = 0.519$ \\
    Per-feature interpretability difference (50M/L27)\DIAG{exp33}{experiments/33\_feature\_interpretability/README.md}; matched at 500M headline (\S\ref{app:interp-500m})\DIAG{exp62}{experiments/62\_interp\_causal\_headline/README.md} & $p = 0.88$ (n.s.) \\
    Sparse probing top-1 $+14.6\%$ (per-feature), $+14.1\%$ (global)\DIAG{exp40/42c/44}{experiments/40\_saprmarks\_recipe/README.md}\DIAG{exp61}{experiments/61\_multiseed\_500m/README.md} & $n = 3$ seeds at 500M (Table~\ref{tab:multiseed-500m}) \\
    \bottomrule
  \end{tabular}
\end{table}

\begin{table}[H]
  \centering
  \small
  \caption{Multi-seed reproducibility at the 500M headline (Qwen3-8B L18, $d_{\mathrm{sae}} = 65{,}536$). Mean $\pm$ SD over three SAE-training seeds $\{42, 123, 456\}$; seed $42$ is the run reported in Table~\ref{tab:headline}.\DIAG{exp61}{experiments/61\_multiseed\_500m/README.md}}
  \label{tab:multiseed-500m}
  \begin{tabular}{lccc}
    \toprule
                  & Standard & Global $a$ & Per-feature \\
    \midrule
    FVE           & $0.7702 \pm 0.0002$ & $0.7690 \pm 0.0000$ & $0.7707 \pm 0.0002$ \\
    Probing top-1 & $0.667 \pm 0.003$ & $0.808 \pm 0.006$ & $0.813 \pm 0.013$ \\
    Top-1 gap     & --- & $+14.1\%$ & $+14.6\%$ \\
    Learned $a$   & --- & $0.258 \pm 0.001$ & $0.076 \pm 0.000$ \\
    \bottomrule
  \end{tabular}
\end{table}

Across three full 500M SAE-training seeds the headline reproduces: FVE matches to $\pm 0.0002$, the top-1 gap is stable at $+14.1$\% (global) and $+14.6$\% (per-feature), and the learned exponent $a$ stays far below the inner-product limit with negligible seed variance. The reproducibility-only Magnitude-Bypass arm and the per-dataset breakdown (Table~\ref{tab:per-dataset-probing}) are reported at the single seed used throughout the main text.

\section{Salient-Concept Control}
\label{app:vignette}

The probing and ablation advantages are aggregate (\S\ref{sec:mechanism}); they do not mean standard scoring fails to represent obvious concepts. To check this directly, we rank each 500M dictionary's features by selectivity for a content concept (mean activation on concept-positive minus concept-negative prompts) and ablate the top-$K$ as a set, measuring the within-arm precision (on-concept effect / off-concept effect) as a function of $K$.\DIAG{exp65}{experiments/65\_causal\_vignette/README.md} For both salient concepts tested (Python code, French text), \emph{both} architectures have a top concept feature that fires strongly on-concept and is silent off-concept (off-concept ablation KL $\approx 0$ for $K \leq 10$ under either score), so neither dictionary lacks the concept. The cosine advantage is therefore not visible at the single-salient-concept level; it emerges in the aggregate over the eight SAEBench probing categories and the long feature tail, consistent with the matched-feature decomposition (\S\ref{sec:mechanism}) and the aggregate ablation-precision result (Table~\ref{tab:tpp-precision}).